\documentclass[journal]{IEEEtran}
\usepackage{cite}

\ifCLASSINFOpdf
   \usepackage[pdftex]{graphicx}
  \graphicspath{{./Figures_SimplePotentialFunction/}}
  \DeclareGraphicsExtensions{.pdf,.jpeg,.png}
\else
  \usepackage[dvips]{graphicx}
  \usepackage{amsmath}
  \graphicspath{{./figures/}}
  \DeclareGraphicsExtensions{.eps}
\fi
\usepackage{tikz}
\usepackage{widetext}
\usepackage{comment}

\usepackage{amsthm,amsmath,amsfonts,dsfont}
\usepackage{enumitem}

\usepackage[linesnumbered,lined,ruled,norelsize]{algorithm2e}
\usepackage{algorithmic}
\SetAlFnt{\small}

\usepackage{array}

\ifCLASSOPTIONcompsoc
  \usepackage[caption=false,font=normalsize,labelfont=sf,textfont=sf]{subfig}
\else
  \usepackage[caption=false,font=footnotesize]{subfig}
\fi
\usepackage{dblfloatfix}

\ifCLASSOPTIONcaptionsoff
  \usepackage[nomarkers]{endfloat}
 \let\MYoriglatexcaption\caption
 \renewcommand{\caption}[2][\relax]{\MYoriglatexcaption[#2]{#2}}
\fi
\usepackage{url}

\makeatletter
\newcommand{\removelatexerror}{\let\@latex@error\@gobble}
\makeatother

\begin{document}
%
\title{Learning Traffic Flow Dynamics using Random Fields}
%
%
%

\author{Saif Eddin Jabari$^{\dagger}$,
		Deepthi Mary Dilip, 
        DianChao Lin, 
        and Bilal Thonnam Thodi
\thanks{$^{\dagger}$Corresponding author. S.E. Jabari is with the Division of Engineering, New York University Abu Dhabi, Saadiyat Island, P.O. Box 129188, Abu Dhabi, U.A.E. and the Department of Civil and Urban Engineering, NYU Tandon School of Engineering, Brooklyn, NY 11201, U.S.A. (e-mail: sej7@nyu.edu)}
\thanks{D. Dilip is with the Department of Civil Engineering, Birla Institute of Technology \& Science, Pilani, Dubai Campus, Dubai International Academic City, Dubai, U.A.E. 
}
\thanks{D. Lin is with the Department of Civil and Urban Engineering, NYU Tandon School of Engineering, Brooklyn, NY 11201, U.S.A. 
}
\thanks{B. Thonnam Thodi is with the Department of Civil and Urban Engineering, NYU Tandon School of Engineering, Brooklyn, NY 11201, U.S.A. 
}
}
\maketitle

\begin{abstract}
This paper presents a mesoscopic traffic flow model that explicitly describes the spatio-temporal evolution of the probability distributions of vehicle trajectories.  The dynamics are represented by a sequence of factor graphs, which enable learning of traffic dynamics from limited Lagrangian measurements using an efficient message passing technique. The approach ensures that estimated speeds and traffic densities are non-negative with probability one. 
The estimation technique is tested using vehicle trajectory datasets generated using an independent microscopic traffic simulator and is shown to efficiently reproduce traffic conditions with probe vehicle penetration levels as little as 10\%. The proposed algorithm is also compared with state-of-the-art traffic state estimation techniques developed for the same purpose and it is shown that the proposed approach can outperform the state-of-the-art techniques in terms reconstruction accuracy.
\end{abstract}


%
\IEEEpeerreviewmaketitle

\section{Introduction}
%
%
%
%

\IEEEPARstart{A}{s} automated vehicle (AV) technologies begin to penetrate vehicle fleets in cities throughout the world, it is reasonable to expect that vehicle trajectory data will become a prominent source of high-resolution traffic data.  AVs may act as probes in the traffic stream, continuously broadcasting their positions and speeds in real-time. More importantly, AVs can also provide distance headways (spacing between successive vehicles) using infrared or radio
technology \cite{yuan2012real,seo2015probe}.  However, privacy issues and technology limitations can limit the ability of traffic management agencies to collect, analyze, and disseminate such information. To overcome this, these data can be 
fused with data obtained from traditional monitoring devices such as inductive-loop detectors (stationary sensors). As these two data sources complement each other, comprehensive datasets can be obtained for traffic monitoring and state estimation \cite{hofleitner2012learning}.  However, the improvement in  accuracy with data fusion, over single sensor applications depends on probe penetration rates and on traffic conditions. In urban road networks, where stationary sensor instrumentation is usually limited and traffic lights play a governing role in the traffic dynamics, a higher number of probes may be necessary to accurately characterize traffic conditions. 

A number of modeling techniques have been proposed in the recent years to estimate traffic densities \cite{herrera2010incorporation,jabari2012stochastic,jabari2013stochastic,deng2013traffic}, speeds \cite{work2008ensemble} and travel times \cite{hellinga2008decomposing,chen2012prediction}.  Studies have also been carried out to extract patterns from streaming data using data mining techniques \cite{hunter2012large, jenelius2017urban,dilip2017sparse,jabariGenGammaKernels}. To account for the variability in urban traffic, a statistical approach using Coupled Hidden Markov Models was proposed in \cite{herring2010estimating} to estimate the traffic state from sparse probe data. The limitations of purely statistical approaches were overcome by \cite{hofleitner2012arterial}, where a hybrid modeling framework combining machine learning with hydrodynamic traffic theory was proposed to predict arterial travel times from streaming GPS probe data. On the other hand, a data-driven model that captures longitudinal interactions between vehicles was proposed in \cite{papathanasopoulou2015towards}.

Research on traffic state estimation (TSE) from probe data for urban networks has mostly focused on the reconstruction of traffic conditions at an aggregate level (over an entire intersection-to-intersection road segment) \cite{furtlehner2007belief,bekiaris2016highway,fountoulakis2017highway}. At a finer scale, traffic densities on a freeway section were reconstructed by modifying traditional continuum models with a correction term to nudge the model estimate towards the GPS probe measurements in \cite{herrera2010incorporation}.  The techniques proposed did not require the knowledge of on- and off-ramp sensor data for density estimation, and minimal penetration rates required on arterial roads was also analyzed. 
 The maximum sampling interval (time between two consecutive probe vehicle samples) required to accurately detect incidents is discussed in \cite{vandenberghe2012feasibility} and the optimal placement of sensors for reliable time-to-detection of incidents is studied in \cite{jabari2016Sensor}.  Comparisons of travel time estimates produced using one source of data versus fused data from two data sources (stationary sensors and probe vehicle data) were performed in \cite{mazare2012trade}. In \cite{kerner2013traffic}, a stochastic model was developed using the three-phase traffic theory to reconstruct the spatio-temporal traffic dynamics using probe vehicle data and fixed detectors located at about 1 km intervals. A number of studies have reported probe penetration rates required for TSE on arterials \cite{hiribarren2014real,ban2009delay,ban2011real,zheng2018traffic,dilip2018vehicle}.  While the reliability of probe vehicle data has been investigated and compared against stationary sensor data \cite{kim2014comparing, bar2007evaluation}, the effect of the variability in the spatio-temporal coverage of probe vehicles on the reconstructed traffic pattern addressed in \cite{kerner2013traffic} needs to be explored further.
 
This paper first develops a probabilistic traffic flow model capable of reproducing shockwaves and the spontaneous emergence of stop-and-go waves in traffic, consistent with empirical observations \cite{sugiyama2008traffic}.  A unique feature of the traffic model, in contrast to all existing models of traffic flow that were developed for TSE, is that it explicitly models the evolution of the probability distribution of vehicle speeds and positions (in a discrete space context). This allows us to ensure that negative traffic states (e.g., speeds and traffic densities) never occur (with probability 1) in our model.   More importantly, this feature is inherited in the proposed TSE method.  To the best of our knowledge this has yet not been addressed in the literature.  For TSE, we first demonstrate that the traffic model possess the structure of Markov random field (MRF).  We then utilize message passing techniques for probabilistic inference of the traffic states with limited measurements.  The main insight is to break the MRF into a sequence of factor graphs, one per time step. The resulting graphs have a tree structure, which allows for exact inference with a single forward pass and a single backward pass; that is, the message passing scheme is computationally efficient.  Moreover, this allows for recursive and online implementations of our algorithm.

The remainder of the paper is organized as follows:  The probabilistic model of mesoscopic traffic dynamics is derived and the inference problem is described next.  This is followed by a representation of the dynamics as a Markov random field the factor graph approach proposed for solving the inference problem.  Model testing and validation experiments follow and a brief summary and discussion of future research conclude the paper.

\section{Stochastic Traffic Dynamics}
\label{S:stochModel}

\subsection{Mesoscopic Flow Model}

A discrete (state and time) \emph{mesoscopic} stochastic model is used to represent the traffic dynamics. Vehicle movement in the model is governed by potential functions that describe the (``energy profile'' of) local traffic conditions.  Similar models have been employed to study interesting traffic phenomena like \emph{synchronized traffic} at ramps and \emph{stop-and-go} regimes \cite{sopasakis2006stochastic}.  This renders them richer than macroscopic models, which capture wave propagation, but not as rich as microscopic models, which capture the motion of individual vehicles in continuous space and time.

The physical roadway is modeled as a one-dimensional uniform lattice $\mathcal{L}$ consisting of $L \equiv |\mathcal{L}|$ cells.  The spatial co-ordinates of each vehicle $n$ on the roadway is discretized in such a way that each cell can be occupied by at most one vehicle, which is achieved by setting the cell length to an appropriate value, e.g. 7.5 m \cite{larraga2005cellular}. The state of each occupied cell at a discrete time $k$ is completely specified by a discretized speed denoted $V_{n}^k \in \{0, \hdots,  v_{\max}\}$, where $v_{\max}$ is the maximum number of cells that can be traversed by a vehicle in one time step.  Clearly, $v_{\max}$ depends on the length of the discrete time step. Thus, a state parameter $\sigma^k_l \in \{-1,0, \hdots,v_{\max}\}$ can be defined for each cell $l \in \mathcal{L}$ at time $k$ to represent the traffic state in the cell, where -1 represents a free (or empty) cell.  We shall also use the vector notation $\mathbf{V}^k \equiv [V_1^k ~ \cdots ~ V_{|\mathcal{V}|}^k]^{\top}$ and $\boldsymbol{\sigma}^k \equiv [\sigma_1^k ~ \cdots ~ \sigma_{L}^k]^{\top}$, where by necessity $|\mathcal{V}| \le L$ (there is at most one vehicle in each cell).

Denote by $\mathcal{V}$ the set of vehicles on the road section and let $n, n-1 \in \mathcal{V}$ be a follower-leader vehicle pair. Let $S_{n}^k$ denote the position of vehicle $n$ in time step $k$ and $G_{n,n-1}^k \equiv |S_{n}^k - S_{n-1}^k|$ the spacing between the two vehicles.  For $v,w \in \{1, \hdots, v_{\max} \}$, define the \emph{interaction potential}
\begin{multline}
	\varepsilon_{n,n-1}^k(v,w \big| \textbf{S}^k) \equiv \frac{1}{\theta_0} \Big(\big\| v - \mathsf{V}(G_{n,n-1}^k) \big\|_2^2 \\+ \big\| w - \mathsf{V}(G_{n-1,n-2}^k) \big\|_2^2\Big),
\label{eqn:int-pot}
\end{multline}
where
\begin{equation}
	\mathsf{V}(G_{n,n-1}^k) = \min \left\{ \Big[ \frac{G_{n,n-1}^k-\theta_1}{\theta_2}  \Big]^{+}, v_{\max} \right\}
\label{eqn:speed-space}
\end{equation}
is a speed-spacing relation and $\textbf{S}^{k} \equiv [S_1^k ~ \cdots ~ S_{|\mathcal{V}|}^k]^{\top}$ is a vector of vehicle positions at time $k$. The projection operator is denoted by $[\cdot]^+ \equiv \max\{\cdot,0\}$. The model parameters are $\theta_0$, $\theta_1$, $\theta_2$ and $v_{\max}$, where $\theta_0$, $\theta_1$ and $\theta_2$ collectively capture driver reaction time and maximum acceleration. The interaction potential for $n = 1$ depends on the boundary conditions.  For example, for a free boundary, one has  
\begin{equation}
	\varepsilon_{1,0}^k(v,w \big| \textbf{S}^k) \equiv \frac{1}{\theta_0} \Big(\big\| v - v_{\max} \big\|_2^2 + \big\| w - v_{\max} \big\|_2^2 \Big).
\end{equation}
For periodic boundary conditions (e.g., a ring road), one has for $n = 1$ that $n-1 = |\mathcal{V}|$.  Note the symmetry $\varepsilon_{n,n-1}^k(v,w | \textbf{S}^k) = \varepsilon_{n-1,n}^k(w,v | \textbf{S}^k)$.
The position of vehicle $n$ at time step $k$ is updated based on its speed $V_{n}^k$ as follows:
\begin{equation}
	S_{n}^{k+1} = \min \big\{ S_{n}^k + V_{n}^k , ~ S_{n-1}^k - 1 \big\}.
\label{eqn:pos-update}
\end{equation} 
The minimum above ensures that vehicles and their leaders do not occupy the same position \cite{wagner1997realistic,li2006realistic}.  Consequently, the state update calculation proceeds in ascending order of $n$ (from the position-wise upstream-most vehicle to the downstream-most vehicle). This is similar to the implicit propagation rule in \cite{emmerich1997improved}, which considers the anticipation of the driver to the movement of the leader vehicle. The traffic state update is carried out in discrete time steps to determine $\boldsymbol{\sigma}^{k+1}$. The position update \eqref{eqn:pos-update} can also be interpreted as a \emph{dynamical potential}
\begin{equation}
	\varepsilon_{n}^k(s \big| \mathbf{S}^{k-1}, \boldsymbol{\sigma}^{k-1}) = \mathds{1}_{\{ s =  \min \{ S_{n}^{k-1} + V_{n}^{k-1} , ~ S_{n-1}^{k-1} - 1 \}\}},
\label{eqn:detNodePot}
\end{equation}
for all $s \in \mathcal{L}$. This results in a deterministic position update rule, given the past. An alternative dynamical potential is given, for any $s \in \mathcal{L}$, by
\begin{equation}
	\varepsilon_{n}^k(s \big| \mathbf{S}^{k-1}, \boldsymbol{\sigma}^{k-1}) = \beta \mathds{1}_{\{ s \le S_{n-1}^{k-1} - 1 \}} \big\| s - S_{n}^{k-1} - V_{n}^{k-1} \big\|_2^2.
\label{eqn:nodePot}
\end{equation}
The parameter $\beta$ in \eqref{eqn:nodePot} is a \emph{precision} parameter; as $\beta$ gets very large \eqref{eqn:nodePot} approaches \eqref{eqn:detNodePot}, while a small $\beta$ corresponds to higher variability.

Since the positions in time step $k$ depend on the states (both positions and speeds) in time step $k-1$ and the speeds depend on the positions  in time step $k$, we can define a state vector-pair $\textbf{Y}^k \equiv \{ \mathbf{S}^k, \boldsymbol{\sigma}^k \}$ and condition both $\varepsilon_{n}^k$ and $\varepsilon_{n,n-1}^k$ on $\textbf{Y}^{k-1}$. The total ``potential energy'' for the $|\mathcal{V}|$ vehicles in the system at time $k$ is
\begin{multline}
	E^{k}(\mathbf{s},\mathbf{v} \big| \textbf{Y}^{k-1}) = \sum_{n \in \mathcal{V}} \Big( \varepsilon_{n}^k(s_{n} \big| \textbf{Y}^{k-1}) \\+ 2 \varepsilon_{n,n-1}^k(v_{n},v_{n-1} \big| \textbf{Y}^{k-1}) \Big).
\label{eqn:energy}
\end{multline}
The probability that the $|\mathcal{V}|$ vehicles assume the speeds $\mathbf{v} \equiv [v_1 \cdots v_{|\mathcal{V}|}]^{\top}$ in the next time step $k+1$, given the state of the system at time $k$ is related to the total energy as
\begin{equation}
\mathbb{P}(\mathbf{S}^k = \mathbf{s}, \mathbf{V}^k = \mathbf{v} \big| \textbf{Y}^{k-1}) \propto e^{-E^{k}(\mathbf{s},\mathbf{v} \big| \textbf{Y}^{k-1})}
\label{eqn:prob}
\end{equation}
for $v_n \in \{0, \hdots, v_{\max} \}$ and $s_n \in \mathcal{L}$, for all $n \in \mathcal{V}$. For any $v_n \not\in \{0, \hdots, v_{\max} \}$ or $s_n \not\in \mathcal{L}$, $\mathbb{P}(\mathbf{S}^k = \mathbf{s}, \mathbf{V}^k = \mathbf{v} | \textbf{Y}^{k-1}) = 0$.  Thus, by construction, the probability of any unreasonable traffic state (e.g., a speed that is negative or greater than $v_{\max}$) is zero.  The standard practice of adding noise to deterministic dynamics cannot preclude such unreasonable traffic states \cite{jabari2012stochastic}.

The steps involved in simulating the traffic dynamics are summarized in Algorithm \ref{alg:1}, which without loss of generality (i) assumes a free downstream boundary, (ii) uses the deterministic dynamical potential \eqref{eqn:detNodePot}, equivalent to the update rule \eqref{eqn:pos-update}, and (iii) takes advantage of the independence structure that results from the way the total energy is calculated in \eqref{eqn:energy}. Also, note the use of lower-case variables (e.g., $v_{n}^k$ instead of $V_{n}^k$); this is done to emphasize that these are simulated realizations and not random quantities. The algorithm can be easily modified to accommodate downstream restrictions in a way that is similar to the upstream state update (see \cite{jabari2016node} for details on boundary treatments) or a periodic boundary and use dynamical potential \eqref{eqn:nodePot}. Finally, a randomization in the braking is introduced through the probability of slow-down $p_2$ as in \cite{schadschneider2002traffic,tian2009synchronized}.

\begin{algorithm}[ht!]
	\caption{Mesoscopic Simulation}
	\label{alg:1}
	\begin{algorithmic}
		\STATE \textbf{Input:}
		\STATE \Indp No. of lattice sites $L:= |\mathcal{L}|$,   No. of time steps $:=K$, 
		\STATE max speed $:= v_{\max}$, model parameters $:= \{\theta_0, \theta_1, \theta_2 \}$,
		\STATE arrival density $:=p_1$, probability of slow-down $:=p_2$
		\STATE \Indm \textbf{Initialize:}
		\STATE \Indp Initial traffic state $:= \boldsymbol{\sigma}^0$
		\STATE \Indm \textbf{Iterate:}
		\STATE \textbf{For} $k := 1$ to $K$ \textbf{do}	
		\STATE \Indp \textbf{For} $n := 1$ to $|\mathcal{V}|$ \textbf{do}
		\STATE \textbf{Position Update:}
		\STATE \Indp \textbf{If} $s_{n}^k := \min \big\{  s_{n}^{k-1} + v_{n}^{k-1} , ~ s_{n-1}^{k-1} - 1 \big\} > L$ \textbf{then}
		\STATE \Indp $n$ leaves the system: $\mathcal{V} := \mathcal{V} \setminus n$ and renumber
		\STATE \Indm \textbf{Else}
		\STATE \Indp Update traffic state: $\sigma^{k}_{s_{n}^{k}} := v_{n}^{k}$
		\STATE \Indm \textbf{End If}	
		\STATE \Indm \textbf{Speed Update}
		\STATE \Indp \textbf{If} $n == 1$ \textbf{then}
		\STATE \Indp $\mathsf{V}( g_{n,n-1}^k) := v_{\max}$ (free boundary)
		\STATE \Indm \textbf{Else}
		\STATE \Indp $g_{n,n-1}^k := |s_{n}^k - s_{n-1}^k|$
		\STATE $\mathsf{V}( g_{n,n-1}^k) := \min \big\{ \big[ (g_{n,n-1}^k-\theta_1) / \theta_2 \big]^+, v_{\max} \big\}$
		\STATE \Indm \textbf{End If}
		\STATE \textbf{For} $v := 0$ to $v_{\max}$ \textbf{do}
		\STATE \Indp{$\varepsilon_{n,n-1}^{k}(v) := \big( v - \mathsf{V}( g_{n,n-1}^k)\big)^2 / \theta_0$}
		\STATE \Indm \textbf{End For}
		\STATE $Z_{n,n-1} :=  \sum_{v=0}^{v_{\max}} \exp (- \varepsilon_{n,n-1}^k(v))$
		\STATE \textbf{For} $v := 0$ to $v_{\max}$ \textbf{do}
		\STATE \Indp $\pi^k_{n,n-1}(v) := \exp(- \varepsilon_{n,n-1}^k(v)) / Z_{n,n-1}$
		\STATE \Indm \textbf{End For}		
		\STATE Sample $v^k_{n}$ from $[\pi_{n,n-1}^k(0) \; \hdots \; \pi_{n,n-1}^k(v_{\max})]^{\top}$
		\STATE $u_1 \sim \text{Uniform}(0,1)$
		\STATE \textbf{If} $u_1 < p_2$ \textbf{then} 
		\STATE \Indp $v^k_{n} := v^k_{n} - 1$
		\STATE \Indm \textbf{End If}
		\STATE \Indm \textbf{End For}
		\STATE \textbf{Boundary Conditions :}
		\STATE	\Indp $u_2 \sim \text{Uniform}(0,1)$
		\STATE \textbf{If} $u_2 < p_1$ and $\sigma^k_1 = 0$ \textbf{then}
		\STATE \Indp Augment new vehicle: $\mathcal{V} := \mathcal{V} \cup |\mathcal{V}|+1$
		\STATE $s_{|\mathcal{V}|}^k := 1$
		\STATE $v_{|\mathcal{V}|}^k := \min \big\{ \big[ (s_{|\mathcal{V}|-1}^k-\theta_1) / \theta_2 \big]^+, v_{\max} \big\}$
		\STATE \Indm \textbf{End If}
		\STATE \Indm \Indm \textbf{End For}
	\end{algorithmic}
\end{algorithm}

\subsection{Probabilistic Inference}
\label{SS:inferenceProb}
Assume there are $|\mathcal{V}|$ vehicles in the system and at each time step $k$, the state (speed and position) of a \textit{subset} of these vehicles is observed.  The estimation problem is concerned with determining the state of all vehicles given the partial observations.  More accurately, the estimation problem seeks to fit the conditional probability distribution of the state given the observations.  Assume the prior distribution of speeds (state at time 0), $\pi_0(\mathbf{s},\mathbf{v}) \equiv \mathbb{P}(\mathbf{S}^0 = \mathbf{s}, \mathbf{V}^0 = \mathbf{v})$, is given.  Let $\boldsymbol{\sigma}^{k,\mathrm{obs}}$ denote the observed traffic states (measurements) available at time $k$.  For each time step, the inference problem seeks to determine the conditional probability
\begin{equation}
	\pi^k(\mathbf{s},\mathbf{v}) \equiv \mathbb{P}\big( \mathbf{S}^k = \mathbf{s}, \mathbf{V}^k = \mathbf{v} \big|\widehat{\mathbf{Y}}^{k-1},\boldsymbol{\sigma}^{k,\mathrm{obs}}\big), \label{Eq_pi}
\end{equation}
where $\widehat{\mathbf{Y}}^{k-1}$ is based on maximum a-posteriori (MAP) estimates of the traffic state in time step $k-1$. Note that we do assume knowledge of the traffic volumes, i.e., the number of vehicles in the system, at each time step.  This assumption is common in the context of TSE techniques based on Lagrangian measurements.  Traffic volumes can be obtained separately, e.g., using the approach outlined in \cite{zheng2017estimating}. 
Inferring the traffic state at time $k$ is achieved using MAP estimates:
\begin{equation}
	\widehat{v}_{n}^k = \underset{v \in \{0,\hdots, v_{\max}\}}{\arg \max} ~ \pi_{n}^k(v) \label{Eq_MAP}
\end{equation}
where
\begin{equation}
	\pi_{n}^k(v)  \equiv \mathbb{P}\big(V_{n}^k = v|\widehat{\mathbf{Y}}^{k-1},\boldsymbol{\sigma}^{k,\mathrm{obs}}\big) \label{Eq_pi1} 
\end{equation}
is the (conditional) marginal probability density of the state of vehicle $n$ at time $k$ given the state at time $k-1$ and observations at time $k$. 
These estimates can then be used to calculate estimated vehicle positions, $\widehat{s}_{n}^k$, via \eqref{eqn:pos-update} and set up $\widehat{\mathbf{Y}}^k$ as input for the next time step.  We break the inference problem into a series of inference problems, one per time step.  The main advantage behind doing this is that the each factor graph (for each time step) is a tree.  This simplifies calculations substantially as demonstrated below.

\section{Factor Graph Representation of the Markov Random Fields}
\label{S:factorGraph}

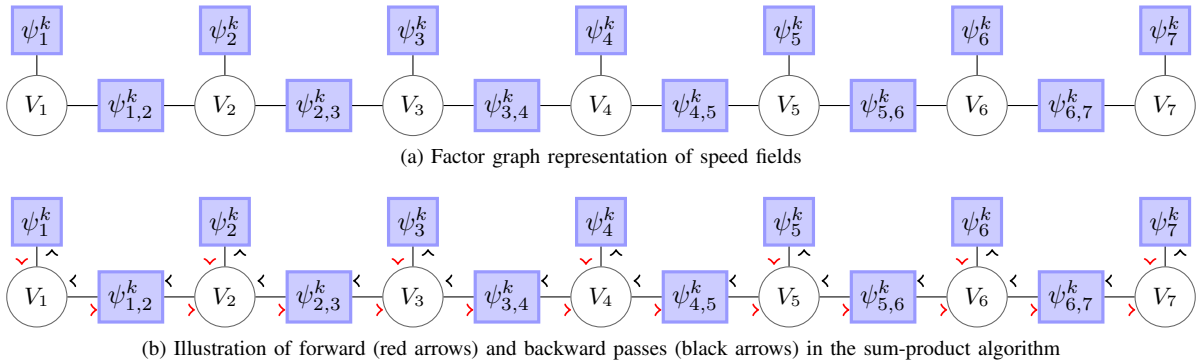
\begin{figure*}[hb!]
	\centering
	\subfloat[][Factor graph representation of speed fields]{
		
		\begin{tikzpicture}[roundnode/.style={circle, draw=black!60,  thin, minimum size=8mm},roundthicknode/.style={circle, draw=red!60, very thick, minimum size=8mm},squarednode/.style={rectangle, draw=blue!40, fill=blue!20, very thick, minimum size=1mm,scale=1.0},][scale=1.0,auto=left,every node/.style={circle,fill=blue!20}]
		\node[squarednode] (f1) at (1,3)  {$\psi_1^k$};
		\node[squarednode] (f2) at (3.5,3)  {$\psi_2^k$};
		\node[squarednode] (f3) at (6,3)  {$\psi_3^k$};
		\node[squarednode] (f4) at (8.5,3)  {$\psi_4^k$};
		\node[squarednode] (f5) at (11,3)  {$\psi_5^k$};
		\node[squarednode] (f6) at (13.5,3)  {$\psi_6^k$};
		\node[squarednode] (f7) at (16,3)  {$\psi_7^k$};
		\node[roundnode] (n1) at (1,2)  {\small $V_1$} ;
		\node[squarednode] (f8) at (2.25,2)  {$\psi_{1,2}^k$};
		\node[roundnode] (n2) at (3.5,2)  {\small $V_2$};
		\node[squarednode] (f9) at (4.75,2) {$\psi_{2,3}^k$};
		\node[roundnode] (n3) at (6,2)  {\small $V_3$};
		\node[squarednode] (f10) at (7.25,2)  {$\psi_{3,4}^k$};
		\node[roundnode] (n4) at (8.5,2) {\small $V_4$};
		\node[squarednode] (f11) at (9.75,2)  {$\psi_{4,5}^k$};
		\node[roundnode] (n5) at (11,2)  {\small $V_5$};
		\node[squarednode] (f12) at (12.25,2) {$\psi_{5,6}^k$};
		\node[roundnode] (n6) at (13.5,2)  {\small $V_6$};
		\node[squarednode] (f13) at (14.75,2)  {$\psi_{6,7}^k$};
		\node[roundnode] (n7) at (16,2)  {\small $V_7$};
		
		Lines
		\draw[-] (n1) -- (f8);
		\draw[-] (f8) -- (n2);
		\draw[-] (n2) -- (f9);
		\draw[-] (f9) -- (n3);
		\foreach \from/\to in {n1/f8,f8/n2,n2/f9,f9/n3,n3/f10,f10/n4,n4/f11,f11/n5,n5/f12,f12/n6,n6/f13,f13/n7,f1/n1,f2/n2,f3/n3,f4/n4,f5/n5,f6/n6,f7/n7}
		\draw (\from) -- (\to);
		\end{tikzpicture}
		
		\label{fig:factor-grapha}
	} 
	
	\subfloat[][Illustration of forward (red arrows) and backward passes (black arrows) in the sum-product algorithm]{
		\begin{tikzpicture}[
		roundnode/.style={circle, draw=black!60,  thin, minimum size=8mm},
		roundthicknode/.style={circle, draw=red!60, very thick, minimum size=8mm},
		squarednode/.style={rectangle, draw=blue!40, fill=blue!20, very thick, minimum size=1mm,scale=1.0},]
		squarednode/.style={rectangle, draw=black!40, fill=gray!15, very thick, minimum size=1mm,scale=1.0},]
		\node[squarednode] (f1) at (1,3)  {$\psi_1^k$};
		\node[squarednode] (f2) at (3.5,3)  {$\psi_2^k$};
		\node[squarednode] (f3) at (6,3)  {$\psi_3^k$};
		\node[squarednode] (f4) at (8.5,3)  {$\psi_4^k$};
		\node[squarednode] (f5) at (11,3)  {$\psi_5^k$};
		\node[squarednode] (f6) at (13.5,3)  {$\psi_6^k$};
		\node[squarednode] (f7) at (16,3)  {$\psi_7^k$};
		\node[roundnode] (n1) at (1,2)  {\small $V_1$} ;
		\node[squarednode] (f8) at (2.25,2)  {$\psi_{1,2}^k$};
		\node[roundnode] (n2) at (3.5,2)  {\small $V_2$};
		\node[squarednode] (f9) at (4.75,2) {$\psi_{2,3}^k$};
		\node[roundnode] (n3) at (6,2)  {\small $V_3$};
		\node[squarednode] (f10) at (7.25,2)  {$\psi_{3,4}^k$};
		\node[roundnode] (n4) at (8.5,2) {\small $V_4$};
		\node[squarednode] (f11) at (9.75,2)  {$\psi_{4,5}^k$};
		\node[roundnode] (n5) at (11,2)  {\small $V_5$};
		\node[squarednode] (f12) at (12.25,2) {$\psi_{5,6}^k$};
		\node[roundnode] (n6) at (13.5,2)  {\small $V_6$};
		\node[squarednode] (f13) at (14.75,2)  {$\psi_{6,7}^k$};
		\node[roundnode] (n7) at (16,2)  {\small $V_7$};
		\draw[red, thick, ->] 
		-- ++(1.1,52pt) -- ++(20pt,0pt) ; 
		\draw[red, thick, ->] 
		-- ++(2.4,52pt) -- ++(20pt,0pt) ; 
		\draw[red, thick, ->] 
		-- ++(3.6,52pt) -- ++(20pt,0pt) ; 
		\draw[red, thick, ->] 
		-- ++(4.9,52pt) -- ++(20pt,0pt) ; 
		\draw[red, thick, ->] 
		-- ++(6.1,52pt) -- ++(20pt,0pt) ; 
		\draw[red, thick, ->] 
		-- ++(7.4,52pt) -- ++(20pt,0pt) ; 
		\draw[red, thick, ->] 
		-- ++(8.6,52pt) -- ++(20pt,0pt) ; 
		\draw[red, thick, ->] 
		-- ++(9.9,52pt) -- ++(20pt,0pt) ; 
		\draw[red, thick, ->] 
		-- ++(11.1,52pt) -- ++(20pt,0pt) ; 
		\draw[red, thick, ->] 
		-- ++(12.4,52pt) -- ++(20pt,0pt) ; 
		\draw[red, thick, ->] 
		-- ++(13.6,52pt) -- ++(20pt,0pt) ; 
		\draw[red, thick, ->] 
		-- ++(14.9,52pt) -- ++(20pt,0pt) ; 
		\draw[red, thick, ->] 
		-- ++(0.8,89pt) -- ++(0pt,-20pt) ; 
		\draw[red, thick, ->] 
		-- ++(3.3,89pt) -- ++(0pt,-20pt) ; 
		\draw[red, thick, ->] 
		-- ++(5.8,89pt) -- ++(0pt,-20pt) ; 
		\draw[red, thick, ->] 
		-- ++(8.3,89pt) -- ++(0pt,-20pt) ; 
		\draw[red, thick, ->] 
		-- ++(10.8,89pt) -- ++(0pt,-20pt) ;
		\draw[red, thick, ->] 
		-- ++(13.3,89pt) -- ++(0pt,-20pt) ; 
		\draw[red, thick, ->] 
		-- ++(15.8,89pt) -- ++(0pt,-20pt) ; 
		\draw[black, thick, ->] 
		-- ++(1.2,55pt) -- ++(0pt,20pt) ;
		\draw[black, thick, ->] 
		-- ++(3.7,55pt) -- ++(0pt,20pt) ;
		\draw[black, thick, ->] 
		-- ++(6.2,55pt) -- ++(0pt,20pt) ;
		\draw[black, thick, ->] 
		-- ++(8.7,55pt) -- ++(0pt,20pt) ;
		\draw[black, thick, ->] 
		-- ++(11.2,55pt) -- ++(0pt,20pt) ;
		\draw[black, thick, ->] 
		-- ++(13.7,55pt) -- ++(0pt,20pt) ;
		\draw[black, thick, ->] 
		-- ++(16.2,55pt) -- ++(0pt,20pt) ;
		\draw[black, thick, ->] 
		-- ++(2.1,63pt) -- ++(-19pt,0pt) ; 
		\draw[black, thick, ->] 
		-- ++(3.3,63pt) -- ++(-17pt,0pt) ; 
		\draw[black, thick, ->] 
		-- ++(4.6,63pt) -- ++(-19pt,0pt) ; 
		\draw[black, thick, ->] 
		-- ++(5.8,63pt) -- ++(-17pt,0pt) ; 
		\draw[black, thick, ->] 
		-- ++(7.1,63pt) -- ++(-19pt,0pt) ; 
		\draw[black, thick, ->] 
		-- ++(8.3,63pt) -- ++(-17pt,0pt) ; 
		\draw[black, thick, ->] 
		-- ++(9.6,63pt) -- ++(-19pt,0pt) ; 
		\draw[black, thick, ->] 
		-- ++(10.8,63pt) -- ++(-17pt,0pt) ; 
		\draw[black, thick, ->] 
		-- ++(12.1,63pt) -- ++(-19pt,0pt) ; 
		\draw[black, thick, ->] 
		-- ++(13.3,63pt) -- ++(-17pt,0pt) ; 
		\draw[black, thick, ->] 
		-- ++(14.6,63pt) -- ++(-19pt,0pt) ; 
		\draw[black, thick, ->] 
		-- ++(15.8,63pt) -- ++(-17pt,0pt) ; 
		
		Lines
		\foreach \from/\to in {n1/f8,f8/n2,n2/f9,f9/n3,n3/f10,f10/n4,n4/f11,f11/n5,n5/f12,f12/n6,n6/f13,f13/n7,f1/n1,f2/n2,f3/n3,f4/n4,f5/n5,f6/n6,f7/n7}
		\draw (\from) -- (\to);
		\end{tikzpicture}
		
		\label{fig:factor-graphb}
	}
	
	\caption{Factor graph and message passing (a.k.a. belief propagation), $V_7$ is the root.}
	\label{fig:factor-graph}
\end{figure*}

\subsection{Traffic Dynamics as Markov Random Fields}
The sequence of vehicle speeds $\mathcal{F} = \{\mathbf{V}^k | k \ge 0 \}$ forms a discrete-valued random field with the Markov property
\begin{equation}
	\mathbb{P}\left( V_{n}^k = v| \mathcal{F} \right)  = \mathbb{P}\left( V_{n}^k = v | \mathcal{F}_{\mathbf{N}_{n}}\right) 
\label{Eq_Markov}
\end{equation}
for all $n \in \mathcal{V}$ and all $k \ge 0$, where $\mathcal{F}_{\mathbf{N}_{n}} \subseteq \mathcal{F}$ is the set of neighbors of $n$, which include both their leader and their follower. The field $\mathcal{F}$ is referred to as a \emph{Markov random field} (MRF).  In each time step $k$, these (conditional) independence assumptions between the variables $V_{n}^k$ can be encoded in a graph $\mathcal{G}^k = (\mathcal{V}^k,\mathcal{E}^k)$  where the set of vehicles $\mathcal{V}^k$ in the system at time $k$ represent the vertices of the graph and the leader-follower interactions between the vehicles in the system at time $k$ represent the edges, $\mathcal{E}^k \in \mathcal{V}^k \times \mathcal{V}^k$, of $\mathcal{G}^k$. 
By encoding the spatial dependencies in the speed field through edges $\mathcal{E}^k$,  the condition in \eqref{Eq_Markov} implies that the speed of any vehicle is independent of the traffic state given the local speed field. That is, if the speeds of the vehicles in $\mathbf{N}_{n}$ are known, then no further knowledge is required to quantify vehicle $n$'s speed probabilities.

\subsection{Factoring and Factor Graph Representation}
The joint probability distribution over all the variables in the Markov model can be compactly represented by defining a set of cliques $\{\mathcal{C}_i\}: \cup_i \mathcal{C}_i = \mathcal{V}^k$, which are subsets of vertices of $\mathcal{G}^k$ such that all vertices in each clique $\mathcal{C}_i$ are completely connected or mutually adjacent. A clique is said to be maximal if no other vertex in $\mathcal{G}^k$ can be added without violating the clique property. 
The joint distribution of the MRF can be expressed over the maximal cliques, as a product of factors as
\begin{equation}
	\pi( \mathbf{v})  = \frac{1}{Z} \prod_{\mathcal{C}_i \subset \mathcal{G}^k} \psi_{\mathcal{C}_i}, \label{Eq_MRF}
\end{equation} 
where $Z$ is the normalizing constant and $\{\psi_{\mathcal{C}_i}\}_{\mathcal{C}_i \subset \mathcal{G}^k}$ is a set of \emph{factors} for each maximal clique $\mathcal{C}_i$ in $\mathcal{G}^k$.  Each factor is a non-negative function defined over a clique to represent the (unnormalized) probability distribution between the vertices in the clique. When the potentials are restricted to be strictly positive, the factors can be re-parametrized in the log space, and expressed in terms of the Boltzmann distribution as $\psi_{\mathcal{C}_i} = e^{-E_{\mathcal{C}_i}(\mathbf{v}_{\mathcal{C}_i})}$, where $ \mathbf{v}_{\mathcal{C}_i}$ is the restriction of $\mathbf{v}$ to the vertices in the clique $\mathcal{C}_i$ and $\{E_{\mathcal{C}_i}(\mathbf{v}_{\mathcal{C}_i})\}_{\mathcal{C}_i \subset \mathcal{G}^k}$ are properly defined potential functions over the cliques. Hence,
\begin{equation}
	\pi( \mathbf{v})  = \frac{1}{Z} e^{-\sum_{\mathcal{C}_i \subset \mathcal{G}^k}E_{\mathcal{C}_i}(\mathbf{v}_{\mathcal{C}_i})}. \label{Eq_MRF2}
\end{equation}
This is analogous to \eqref{eqn:prob}: the lower the energy of the clique configuration (of the states), the higher the probability of the configuration.  Traditional techniques used to represent the dependency in the traffic dynamics graphically would result in what is known as a loopy Markov model.  These are known to not factor uniquely.  Instead, the proposed approach defines a partially directed graph, or a Markov random field, parameterized on a set of factors based on the stochastic traffic dynamics presented above.  Conditioning on $\widehat{\mathbf{Y}}^{k-1}$, $\pi^k(\mathbf{v}) = \prod_{n \in \mathcal{V}^k} \pi_{n}^k(v_{n})$.

Consider the case where probe vehicles are equipped with sensors capable of measuring distances and speeds of other vehicles that are immediately adjacent, their immediate leaders and followers.  (We note that forward and rear collision warning systems, which standard in most present-day vehicles, possess these capabilities.) Dependence on the past is encoded into \emph{node factors}, $\{\psi_{n}^k\}_{n \in \mathcal{V}^k}$, and interactions between adjacent vehicles are encoded into \emph{edge potentials}, $\{\psi_{n_1,n_2}^k\}_{(n_1,n_2) \in \mathcal{E}^k}$. (When two vehicle indices appear in the subscript, the factor is to be implicitly understood as an edge factor.)  
The node factors are related to the traffic dynamics via the dynamical potentials: 
\begin{equation}
	\psi_{n}^k = \big[e^{-\varepsilon_{n}^k(1|\widehat{\mathbf{Y}}^{k-1})} ~ \cdots ~ e^{-\varepsilon_{n}^k(L|\widehat{\mathbf{Y}}^{k-1})}\big]^{\top} 
\end{equation}
and the edge factors are related to the traffic dynamics via the interaction potentials:
\begin{multline}
	\psi_{n_1,n_2}^k \\= \begin{bmatrix}
	e^{-\varepsilon_{n_1,n_2}^k(0,0|\widehat{\mathbf{Y}}^{k-1})} \quad \cdots \quad  e^{-\varepsilon_{n_1,n_2}^k(0,v_{\max}\widehat{\mathbf{Y}}^{k-1})} \\
	 \ddots  \\
	e^{-\varepsilon_{n_1,n_2}^k(v_{\max},0|\widehat{\mathbf{Y}}^{k-1})}  \cdots   e^{-\varepsilon_{n_1,n_2}^k(v_{\max},v_{\max}\widehat{\mathbf{Y}}^{k-1})}
	\end{bmatrix}.
\end{multline}
Fig. \ref{fig:factor-grapha} presents a factor graph representation of a system with five vehicles along with interactions between adjacent vehicles.  The vehicle speeds are represented by circular nodes in the graph. Factors are represented by square nodes in the graph.  

\subsection{Message Passing and Sum-Product Algorithm}

The factor graph framework enables exact inference of the local marginals over nodes or subsets of nodes in tree-structured graphs (such as the one depicted in Fig. \ref{fig:factor-graph}) using the sum-product algorithm (a.k.a. Bayesian belief propagation). The sum-product algorithm computes the marginal distributions of elements of the random field as a product of the incoming \emph{messages} from its neighboring factor nodes, as shown in Fig. \ref{fig:factor-graphb}. A message $\mu_{\psi_{\mathcal{C}_i} \rightarrow V_{n}}$ from a factor $\psi_{\mathcal{C}_i}$ to a variable $V_{n}$ can be interpreted as the information contained in the factor about the variable.  
The tree structure of the factor graph allows for exact inference with a single forward pass and a single backward pass, starting from any node as a root node. The red arrows in Fig. \ref{fig:factor-graphb} illustrate the forward pass when $V_7$ is the root and the black arrows represent the backward pass.  The sum-product algorithm as applied in our context is shown in Algorithm \ref{alg:2}, where we have adopted the convention that the root is always the downstream-most vehicle in the system, $n = |\mathcal{V}^k|$. 
\begin{algorithm}[ht!]
	\caption{Sum-Product Message Passing}
	\label{alg:2}
	\begin{algorithmic}
		\STATE \textbf{For} $k := 1$ to $K$ \textbf{do}	
		\STATE \Indp $\widehat{\mathbf{Y}}^k := \{ \widehat{\mathbf{s}}^k, \widehat{\boldsymbol{\sigma}}^k \}$
		\STATE \textbf{Forward Pass:}
		\STATE \textbf{For} $n := 1$ to $|\mathcal{V}^k|$ \textbf{do}
		\STATE \Indp $\mu_{\psi^k_{n} \rightarrow V^k_{n}} := \psi_{n}^k$
		\STATE \textbf{If} $n == 1$ \textbf{then}
		\STATE \Indp $\mu_{V^k_{n} \rightarrow \psi^k_{n,n+1}} := \mu_{\psi^k_{n} \rightarrow V^k_{n}}$
		\STATE \Indm \textbf{Else}
		\STATE \Indp $\mu_{\psi_{n-1,n}^k \rightarrow V_{n}^k}(v) := \big(\psi_{n-1,n}^k\big)^{\top} \mu_{V_{n-1}^k \rightarrow \psi_{n-1,n}^k}$
		\STATE $\mu_{V^k_{n} \rightarrow \psi^k_{n,n+1}} := \mathsf{diag} \big(\mu_{\psi^k_{n} \rightarrow V^k_{n}} \big) \mu_{\psi^k_{n-1,n} \rightarrow V^k_{n}}$
		\STATE \Indm \textbf{End If}
		\STATE \Indm \textbf{End For}
		\STATE \textbf{Backward Pass:}
		\STATE \textbf{For} $n := |\mathcal{V}^k|$ to $1$ \textbf{do}
		\STATE \Indp \textbf{If} $n == |\mathcal{V}^k|$ \textbf{then}
		\STATE \Indp $\mu_{V_{n}^k \rightarrow \psi_{n}^k} := \mu_{\psi_{n-1,n}^k \rightarrow V_{n}^k}$
		\STATE $\mu_{V_{n}^k \rightarrow \psi_{n-1,n}^k} := \mu_{\psi_{n}^k \rightarrow V_{n}^k}$
		\STATE \Indm \textbf{Else If} $n == 1$ \textbf{then}
		\STATE \Indp $\mu_{\psi_{n,n+1}^k \rightarrow V_{n}^k} := \psi_{n,n+1}^k \mu_{V_{n+1}^k \rightarrow \psi_{n,n+1}^k}$
		\STATE $\mu_{V_{n}^k \rightarrow \psi_{n}^k} := \mu_{\psi_{n,n+1}^k \rightarrow V_{n}^k}$
		\STATE \Indm \textbf{Else}
		\STATE \Indp $\mu_{\psi_{n,n+1}^k \rightarrow V_{n}^k} := \psi_{n,n+1}^k \mu_{V_{n+1}^k \rightarrow \psi_{n,n+1}^k}$
		\STATE $\mu_{V_{n}^k \rightarrow \psi_{n}^k} := \mathsf{diag}\big(\mu_{\psi_{n-1,n}^k \rightarrow V_{n}^k}\big) \mu_{\psi_{n,n+1}^k \rightarrow V_{n}^k}$
		\STATE $\mu_{V_{n}^k \rightarrow \psi_{n-1,n}^k} := \mu_{\psi_{n}^k \rightarrow V_{n}^k} \mu_{\psi_{n,n+1}^k \rightarrow V_{n}^k}$
		\STATE \Indm \textbf{End If}
		\STATE \Indm \textbf{End for}
		\STATE \textbf{Marginal Distributions:} 
		\STATE \textbf{For} $n := 1$ to $|\mathcal{V}^k|$ \textbf{do}
		\STATE \Indp \textbf{If} $n == 1$ \textbf{then}
		\STATE \Indp $\widetilde{\pi}_{n}^k := \mathsf{diag}\big( \mu_{ \psi_{n}^k \rightarrow V_{n}^k } \big) \mu_{ \psi_{n,n+1}^k \rightarrow V_{n}^k }$
		\STATE \Indm \textbf{Else if} $n == |\mathcal{V}^k|$ \textbf{then}
		\STATE \Indp $\widetilde{\pi}_{n}^k := \mathsf{diag}\big( \mu_{ \psi_{n}^k \rightarrow V_{n}^k } \big) \mu_{ \psi_{n-1,n}^k \rightarrow V_{n}^k }$
		\STATE \Indm \textbf{Else}
		\STATE \Indp $\widetilde{\pi}_{n}^k := \mathsf{diag}\big( \mu_{ \psi_{n}^k \rightarrow V_{n}^k } \big) \mathsf{diag}\big(\mu_{ \psi_{n-1,n}^k \rightarrow V_{n}^k } \big) \mu_{ \psi_{n,n+1}^k \rightarrow V_{n}^k }$
		\STATE \Indm \textbf{End If}
		\STATE $\pi_{n}^k := \widetilde{\pi}_{n}^k / \big(\mathbf{1}_v^{\top} \widetilde{\pi}_{n}^k\big)$
		\STATE $\widehat{v}_{n}^{k+1} := \arg \max_{v \in \{0,\hdots, v_{\max}\}} ~ \pi_{n}^k(v)$
		\STATE $\widehat{s}_{n}^{k+1} := \min \big\{ \widehat{s}_{n}^k + \widehat{v}_{n}^k , ~ \widehat{s}_{n-1}^k - 1 \big\}$
		\STATE \Indm \textbf{End for}
		\STATE \textbf{Update Boundaries:}
		\STATE \textbf{If} $\widehat{s}_1^{k+1} > L$ \textbf{then}
		\STATE \Indp Remove veh. 1: $\mathcal{V}^{k+1} := \mathcal{V}^k \setminus 1$ and renumber
		\STATE \Indm \textbf{End If}
		\STATE \textbf{If} new vehicle enters the system \textbf{then}
		\STATE \Indp $\mathcal{V}^{k+1} := \mathcal{V}^{k+1} \cup |\mathcal{V}^{k+1}|+1$ and update state
		\STATE \Indm \textbf{End If}
		\STATE \Indm \textbf{End For}
	\end{algorithmic}
\end{algorithm}
The operation $\mathsf{diag}(\mu)$ creates a diagonal matrix with element given by the vector $\mu$ and $\mathbf{1}_v$ is a vector of ones of size $v_{\max}+1$.  More details on the sum-product algorithm can be found in standard references on graphical models/machine learning; e.g., \cite{bishop2006pattern,koller2009probabilistic}. Algorithm \ref{alg:2} seeks to determine the entire marginal distributions, as opposed to calculating the probabilities of single values.  The following list illustrates how the messages are calculated (we have removed the superscript $k$ to simplify notation):
\begin{enumerate}[wide, labelwidth=!, labelindent=0pt]
	\item From $V_{n}$ to $\psi_{n}$:
	\begin{multline}
		\mu_{V_{n} \rightarrow \psi_{n}}(v) \\= \left\{
		\begin{array}{ll} 
		\mu_{\psi_{n,n+1} \rightarrow V_{n}}(v) &  n = 1 \\
		\mu_{\psi_{n-1,n} \rightarrow V_{n}}(v) \mu_{\psi_{n,n+1} \rightarrow V_{n}}(v) & 1 < n < |\mathcal{V}| \\
		\mu_{\psi_{n-1,n} \rightarrow V_{n}}(v) & n = |\mathcal{V}|
		\end{array}. \right.
	\end{multline}
	\item From $V_{n}$ to $\psi_{n,n + 1}$:
	\begin{multline}
		\mu_{V_{n} \rightarrow \psi_{n,n+1}}(v) \\= \left\{
		\begin{array}{ll} 
		\mu_{\psi_{n} \rightarrow V_{n}}(v) &  n = 1 \\
		\mu_{\psi_{n} \rightarrow V_{n}}(v) \mu_{\psi_{n-1,n} \rightarrow V_{n}}(v) & n \ne 1
		\end{array}. \right.
	\end{multline}
	\item From $V_{n}$ to $\psi_{n-1,n}$:
	\begin{multline}
	\mu_{V_{n} \rightarrow \psi_{n-1,n}}(v) \\= \left\{
	\begin{array}{ll} 
	\mu_{\psi_{n} \rightarrow V_{n}}(v) &  n = |\mathcal{V}| \\
	\mu_{\psi_{n} \rightarrow V_{n}}(v) \mu_{\psi_{n,n+1} \rightarrow V_{n}}(v) & n \ne |\mathcal{V}|
	\end{array}. \right.
	\end{multline}
	\item From $\psi_{n}$ to $V_{n}$:
	\begin{equation}
		\mu_{\psi_{n} \rightarrow V_{n}}(v) = \psi_{n}(v).
	\end{equation}
	\item From $\psi_{n,n+1}$ to $V_{n}$:
	\begin{equation}
		\mu_{\psi_{n,n+1} \rightarrow V_{n}}(v) = \sum_{z=0}^{v_{\max}} \psi_{n,n+1}(v,z) \mu_{V_{n+1} \rightarrow \psi_{n,n+1}}(z).
	\end{equation}
	\item From $\psi_{n-1,n}$ to $V_{n}$:
	\begin{equation}
		\mu_{\psi_{n-1,n} \rightarrow V_{n}}(v) = \sum_{z=0}^{v_{\max}} \psi_{n-1,n}(z,v) \mu_{V_{n-1} \rightarrow \psi_{n-1,n}}(z).
	\end{equation}
\end{enumerate}

\begin{figure*}[ht!]
	\centering	
	\begin{tikzpicture}[
	roundnode/.style={circle, draw=black!60,  very thick, minimum size=8mm},
	roundfillnode/.style={circle, draw=red!60, fill=red!20, very thick, minimum size=8mm},
	squarednode/.style={rectangle, draw=black!40, fill=blue!20, very thick, minimum size=1mm,scale=1.0},
	squaredfillnode/.style={rectangle, draw=red!40, fill=red!20, very thick, minimum size=1mm,scale=1.0}]
	[scale=.8,auto=left,every node/.style={circle,fill=blue!20}]
	
	\node[squarednode] (f1) at (1,3.3)  {$\psi_1^k$};
	\node[squarednode] (f2) at (3.5,3.3)  {$\psi_2^k$};
	\node[squaredfillnode] (f3) at (6,3.3)  {$\psi_3^k$};
	\node[squarednode] (f4) at (8.5,3.3)  {$\psi_4^k$};
	\node[squarednode] (f5) at (11,3.3)  {$\psi_5^k$};
	\node[squaredfillnode] (f6) at (13.5,3.3)  {$\psi_6^k$} ;
	\node[squarednode] (f7) at (16,3.3)  {$\psi_7^k$};
	\node[roundnode] (n1) at (1,2) {\small $V_1$};
	\node[squarednode] (f8) at (2.25,2)  {$\psi_{1,2}^k$};
	\node[roundnode] (n2) at (3.5,2)  {\small $V_2$};
	\node[squaredfillnode] (f9) at (4.75,2) {$\psi_{2,3}^k$};
	\node[roundfillnode] (n3) at (6,2)  {\small $V_3=v_3$};
	\node[squaredfillnode] (f10) at (7.25,2)  {$\psi_{3,4}^k$};
	\node[roundnode] (n4) at (8.5,2) {\small $V_4$};
	\node[squarednode] (f11) at (9.75,2)  {$\psi_{4,5}^k$};
	\node[roundnode] (n5) at (11,2)  {\small $V_5$};
	\node[squaredfillnode] (f12) at (12.25,2) {$\psi_{5,6}^k$};
	\node[roundfillnode] (n6) at (13.5,2)  {\small $V_6=v_6$};
	\node[squaredfillnode] (f13) at (14.75,2)  {$\psi_{6,7}^k$};
	\node[roundnode] (n7) at (16,2)  {\small $V_7$};
	Lines
	
	\draw[-] (n1) -- (f8);

	\foreach \from/\to in {n1/f8,f10/n4,f8/n2,n2/f9,n4/f11,f11/n5,n5/f12,f13/n7,f1/n1,f2/n2,f3/n3,f4/n4,f5/n5,f6/n6,f7/n7}
	\draw (\from) -- (\to);
	
	\end{tikzpicture}
	\caption{Decomposition of the factor graph into a forest of independent sub-graphs in the presence of measurements.}
	\label{fig:factor-graphc}
\end{figure*}
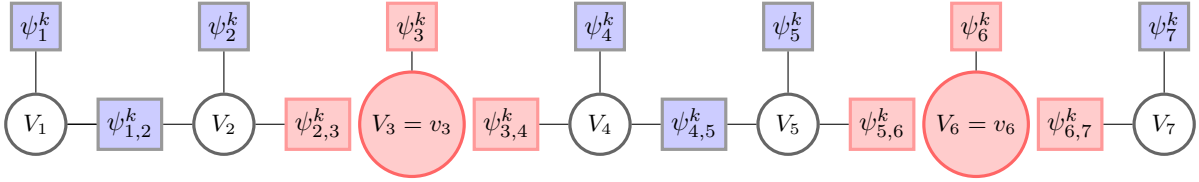 

When measurements are available for some of the vehicles, given by $\sigma^{k,\mathrm{obs}}$, conditioning on such measurements can be done by \emph{clamping} the corresponding variables to the observed values.  As a consequence of conditional independence, when a node is observed, it breaks the chain structure into a forest of independent chains.  For example, for the seven vehicle system in Fig. \ref{fig:factor-grapha}, assuming $V_3=v_3$ and $V_6=v_6$ are given (i.e., the speeds of vehicles 3 and 6 are known/measured), the independent forest shown in Fig. \ref{fig:factor-graphc} is obtained.  

Note how, by design, the only values that the speeds can take in $\psi_{n_1,n_2}^k$ fall in the set $\{0, \hdots, v_{\max}\}$, the positions can only take values in $\mathcal{L}$ in $\psi_n^k$, and the dynamics are collision-free, via the `min' operation. These ensure that speeds are non-negative (and cannot exceed $v_{\max}$) with probability 1. Since message are only products of these potentials, this is preserved in all estimated traffic states obtained with Algorithm \ref{alg:2}.

The inference problem mainly involves performing a single forward pass and a single backward pass, the number of calculations is directly proportional to $\sum_{k=1}^K(5|\mathcal{V}^k|-2) < (5L-2)K$.  The average number of calculations involved in computing the messages and the probabilities is bounded from above by $C \equiv (v_{\max}+1)^2$. This (loosely) results in a computational complexity of $O(CLK)$. 

\section{Model Testing and Validation}
\label{S:numerics}
\subsection{Illustration of Traffic Dynamics}
\begin{figure}[h!]
	\centering
	\subfloat[][Normal traffic conditions]{
		\includegraphics[width=0.4\textwidth]{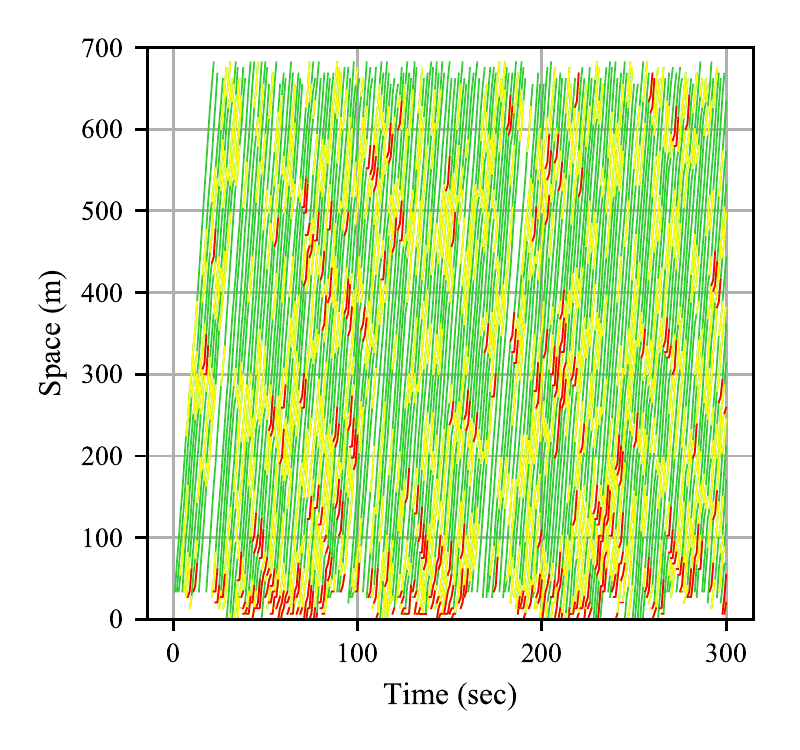}
		\label{fig:long-vta}} 

	\subfloat[][Traffic incident for $20$sec]{
		\includegraphics[width=0.4\textwidth]{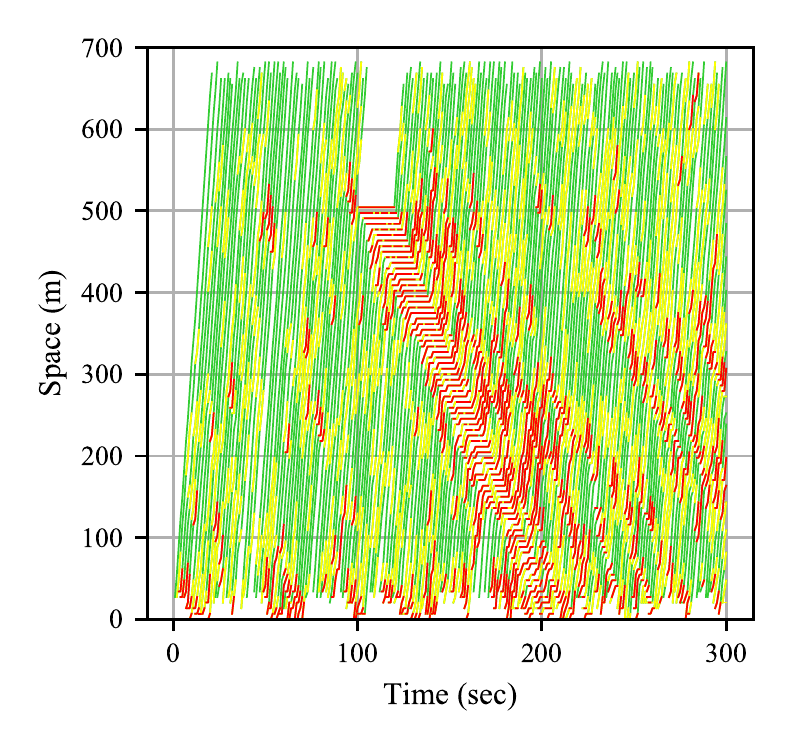}
		\label{fig:long-vtb}}

	\caption{Simulated vehicle trajectories on a $700$m long road section with arrival density, $p_1=0.7$. Color coding represents the speed ranges. \emph{Green} - free flow $(70-120\text{kmph})$, \emph{yellow} - super critical traffic densities $(25-70\text{kmph})$ and \emph{red} - congested traffic $(0-25\text{kmph})$.}
	\label{fig:long-vt}
\end{figure}

\begin{figure}[h!]
	\centering
	\subfloat[][ 20 vehicles uniformly spaced at time zero]{
		\includegraphics[width=0.4\textwidth]{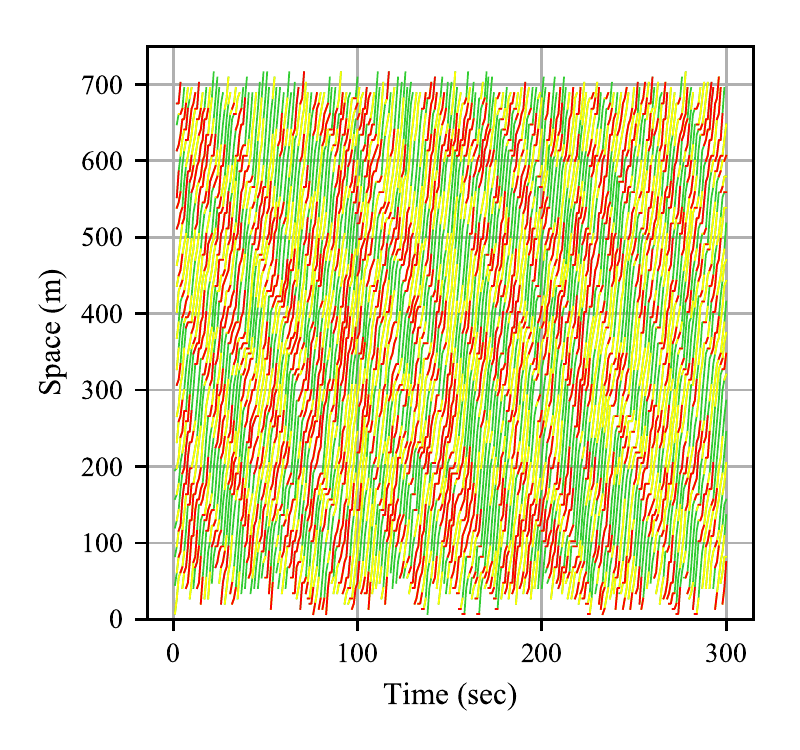}
		\label{fig:ring-vta}} 
	
	\subfloat[][30 vehicles uniformly spaced at time zero]{
		\includegraphics[width=0.4\textwidth]{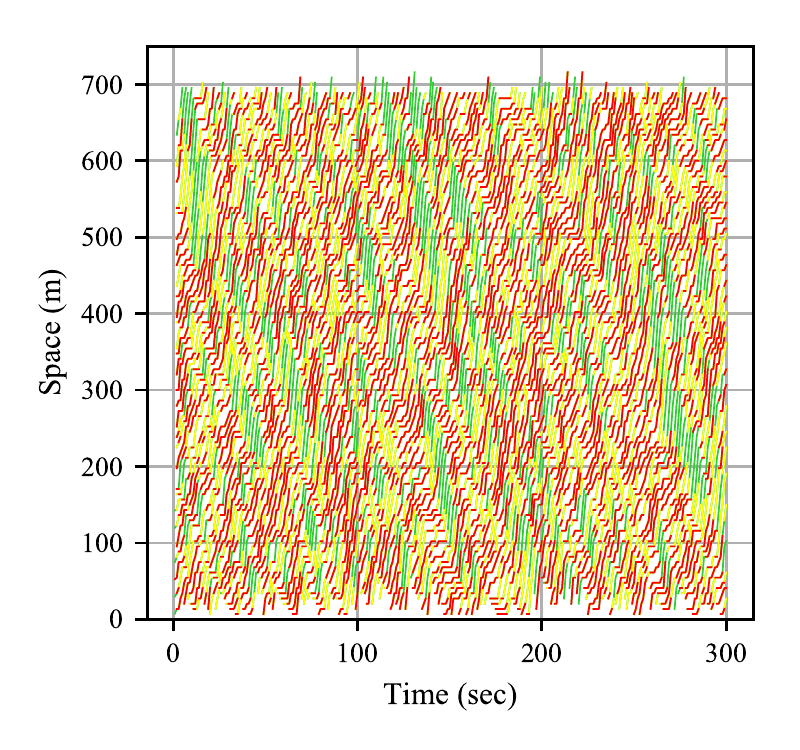}
		\label{fig:ring-vtb}}

	\caption{Simulated vehicle trajectories on a 700m length ring road with different numbers of vehicles. Color coding represents the speed ranges, \emph{Green} - free flow $(70-120\text{kmph})$, \emph{yellow} - super critical traffic densities $(25-70\text{kmph})$ and \emph{red} - congested traffic $(0-25\text{kmph})$.}
	\label{fig:ring-vt}
\end{figure}

We first illustrate the properties of the traffic dynamics obtained with Algorithm \ref{alg:1}.  To do, we simulate different initial and boundary conditions, and investigate the emerging vehicle trajectories and the fundamental diagram. We consider two scenarios: (a) a \emph{long road} section and (b) a \emph{ring road}, each of which has a total length of 700m. We assume the following model parameters: $v_{\max}=34$m/sec, $\theta_0=110\mathrm{m}^2/\mathrm{sec}^2$, $\theta_1=7.5$m, and $\theta_2=1.1$sec. The generated vehicle trajectories are plotted in Fig.\ref{fig:long-vt} and Fig.\ref{fig:ring-vt}. The trajectories are color coded based on vehicle speeds, \emph{green} for free flow states (70-120kmph), \emph{yellow} for super-critical traffic states (25-70kmph) and \emph{red} for heavily congested states (0-25kmph).

Figures \ref{fig:long-vta} and \ref{fig:long-vtb} show the vehicle trajectories along the long road section under normal traffic conditions and incident traffic (a stopped vehicle), respectively. In Fig. \ref{fig:long-vta}, one can observe the onset and dissipation of minor shockwave patterns caused by the high density traffic. This indicates that vehicles adapt their speed to local traffic conditions. One can also observe the backward propagating shockwave caused by the traffic incident in Fig. \ref{fig:long-vtb}. The speed of the backward propagating shockwave is approximately $21$kmph, which is reasonable for freeway traffic \cite{lu2007freeway}. Also, the traffic queue dissipates at a much faster rate than the queue builds up, implying maximum saturation flow when clearing the queue.

Figures \ref{fig:ring-vta} and \ref{fig:ring-vtb} show the vehicle trajectories obtained for a ring road 20 and 30 vehicles, respectively. In both cases, the vehicles are uniformly distributed along the road at time zero. By closely observing the red patterns in both figures, one observes the formation stop-and-go waves consistent with empirical observations \cite{sugiyama2008traffic}. One also observes that stop-and-go waves occur with higher frequency (as to be expected) in Fig. \ref{fig:ring-vtb}.

\begin{figure}[h!]
	\centering
	\subfloat[][ Flow-density diagram]{
		\includegraphics[width=0.4\textwidth]{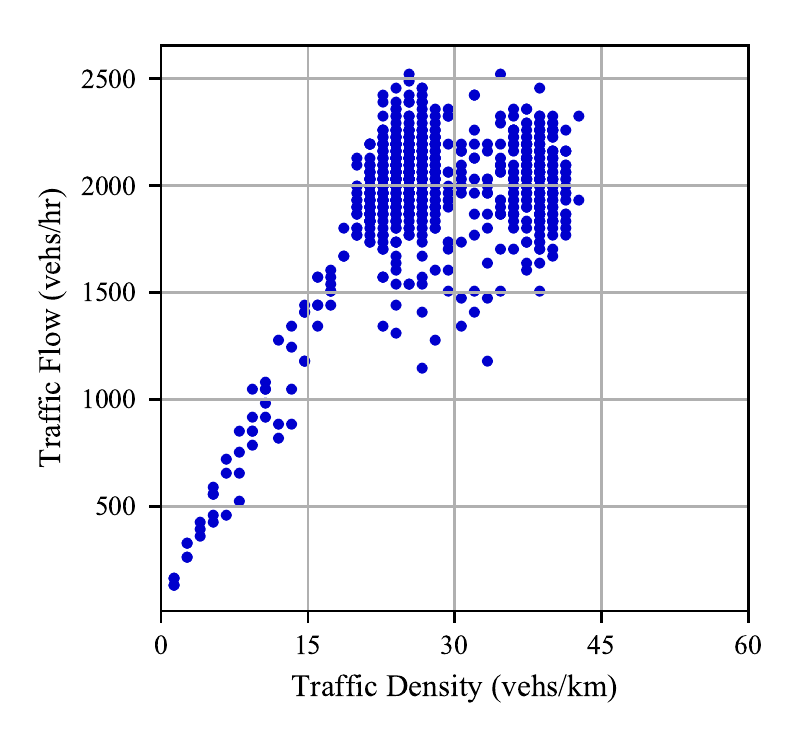}
		\label{fig:fund-diaga}} 
			
	\subfloat[][Speed-density diagram]{
		\includegraphics[width=0.4\textwidth]{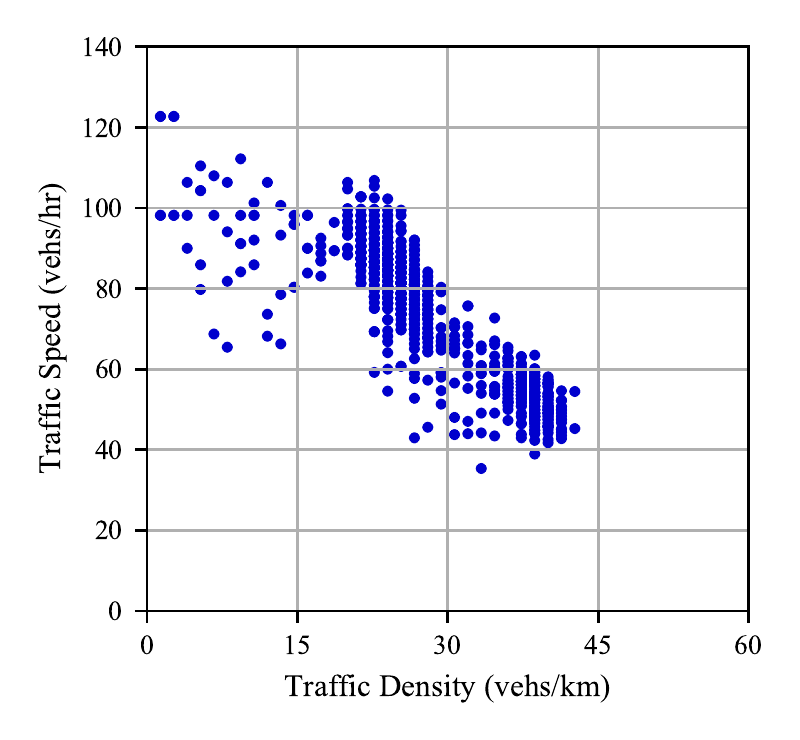}
		\label{fig:fund-diagb}}

	\caption{Fundamental diagram computed for the long road section.}
	\label{fig:fund-diag}
\end{figure}

Figures \ref{fig:fund-diaga} and \ref{fig:fund-diagb} are scatterplots of flow-density and speed-density data, respectively, extracted from the the long road section example. We see a critical density around critical density 25vehs/km in Fig.  \ref{fig:fund-diaga} and a clear difference between free-flowing traffic and (moderately) congested traffic.  We see less scatter in the former and heavier scatter in the latter, consistent with typical field-observed scatterplots obtained along freeways.

\subsection{Reconstruction using MRF: Simulation Example}

{ 
In this section, we perform simulation experiments using a commercial microscopic traffic simulator to generate ``ground truth'' dynamics.  Samples of simulated vehicle trajectories are then used to reconstruct all vehicle dynamics using Algorithm \ref{alg:2}.  We simulate a freeway section with a free-flow speed of 110km/hr with an on-ramp located at the downstream end of the road.  Flow through the on-ramp creates stop-and-go waves along the freeway segment.  Fig. \ref{fig:actual} provides a speed map of the simulated ``ground truth'' dynamics.

\begin{figure}[ht!]%
	\centering
	
	\resizebox{0.45\textwidth}{!}{%
		\includegraphics{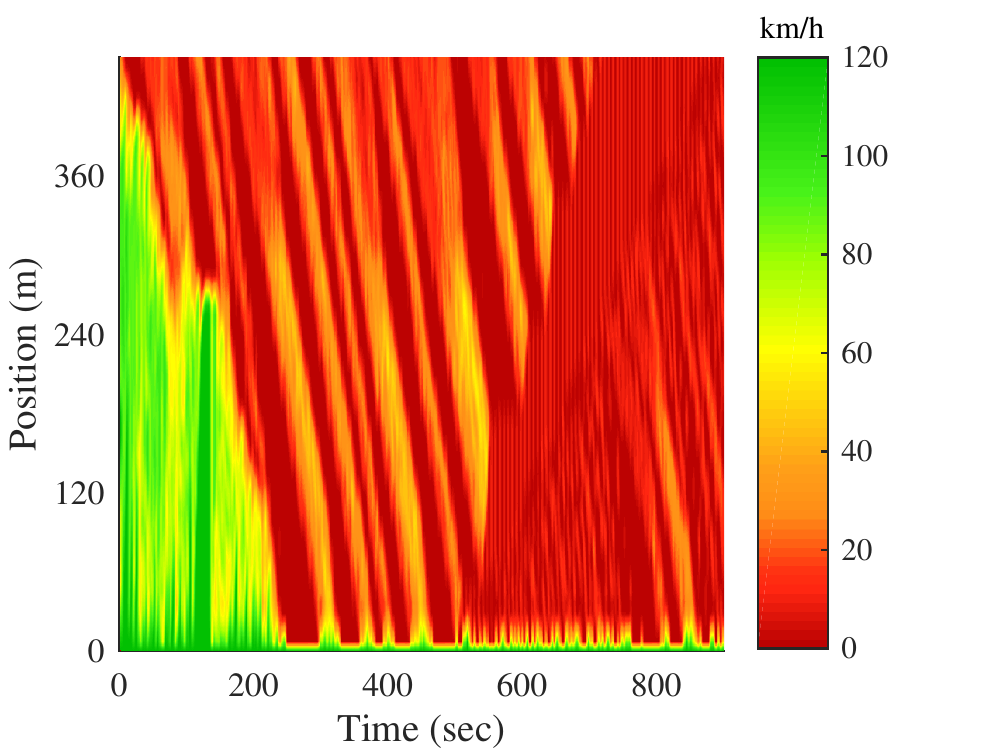}}
	
	\caption{Simulated ``ground truth'' speed map.  }%
	\label{fig:actual}%

\end{figure}

To test the reconstruction algorithm, we set up a lattice of cells of length 7.5m each with $v_{\max} = 4$ (corresponding to 108km/hr).  We compare the reconstructed speed map using Algorithm \ref{alg:2} under varying levels of probe vehicle penetration.  

\begin{figure}[h!]
	\centering
	\subfloat[][Probe penetration =10\%. $\epsilon_{\sigma}^{\mathrm{rel}} = 0.2876$, RMSE = 18.25 km/hr]{
		\includegraphics[width=0.45\textwidth]{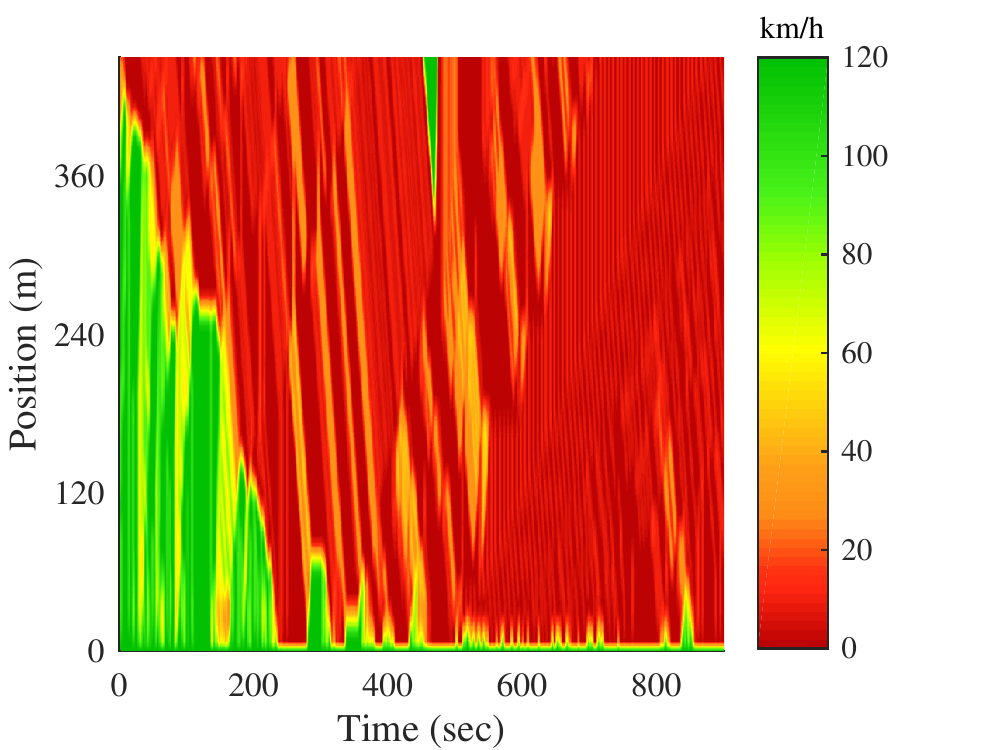}
		\label{fig:estimateda}} 
	
	\subfloat[][Probe penetration =15\%. $\epsilon_{\sigma}^{\mathrm{rel}} = 0.2757$, RMSE =17.49kmph]{
		\includegraphics[width=0.45\textwidth]{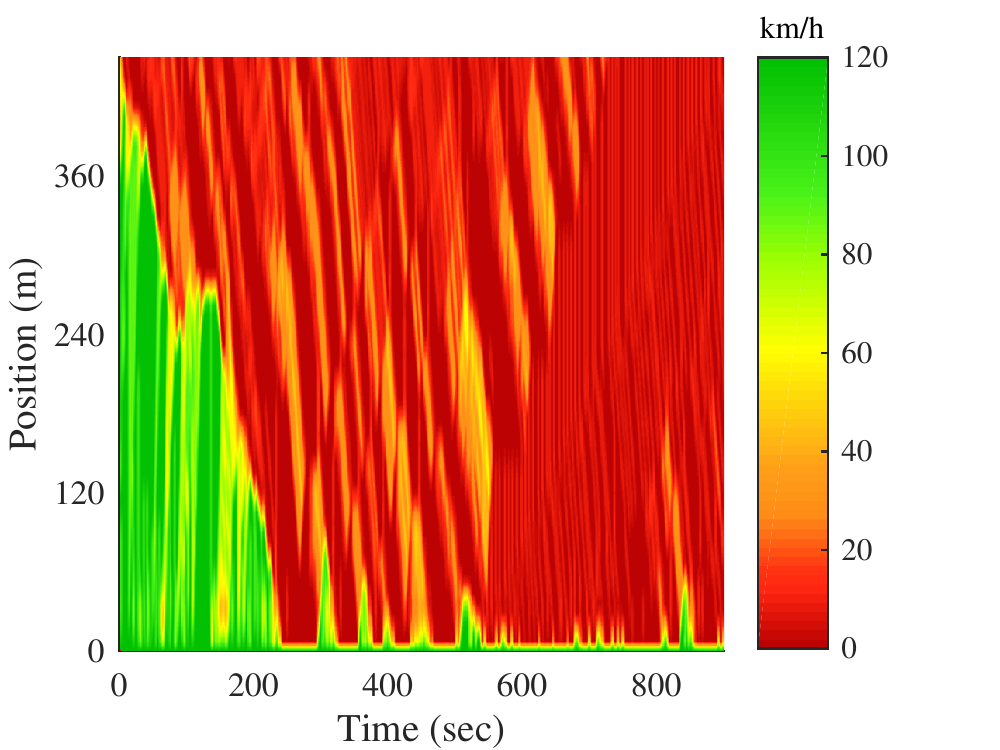}
		\label{fig:estimatedb}}
	
	\subfloat[][Probe penetration =20\%. $\epsilon_{\sigma}^{\mathrm{rel}} = 0.1926$, RMSE = 12.22kmph]{
		\includegraphics[width=0.45\textwidth]{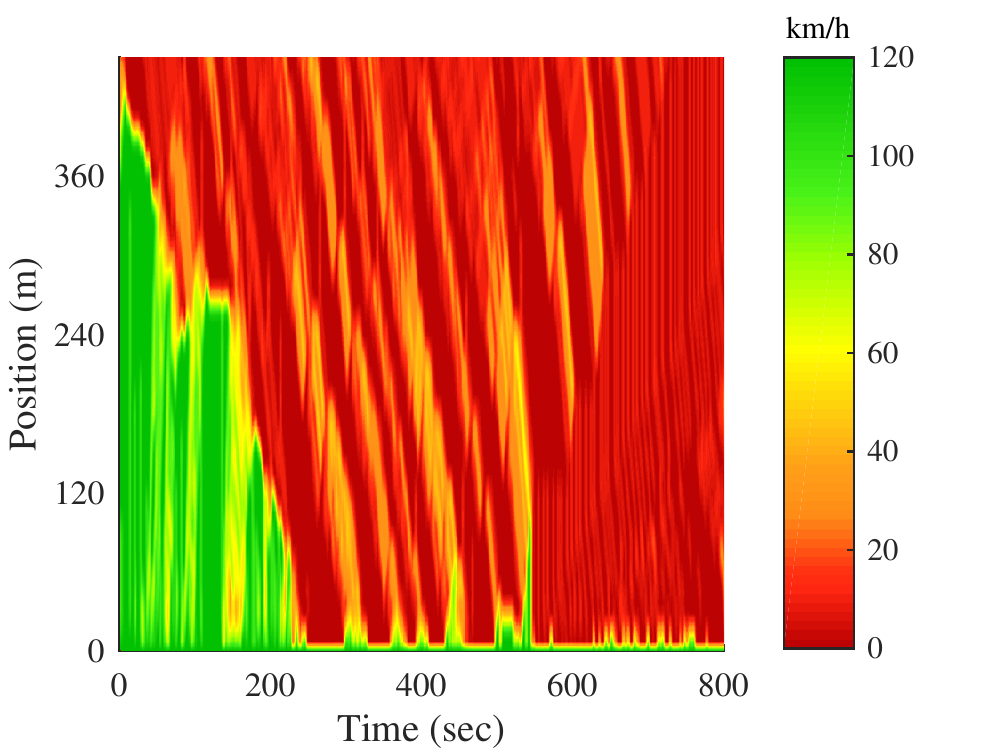}
		\label{fig:estimatedc}}

	\caption{Estimated spatio-temporal speed maps at different probe penetrations.}
	\label{fig:estimated}
\end{figure}

The effect of increasing the probe penetration rate on the reconstructed speed maps is illustrated in Fig. \ref{fig:estimated}, which we also compare with the ground truth speeds using root mean square errors (RMSEs) and the relative error in speed state, $\epsilon_{\sigma}^{\mathrm{rel}}$:
\begin{equation}
\epsilon_{\sigma}^{\mathrm{rel}} = \frac{\sqrt{KL\sum_{k=1}^K \sum_{l=1}^L(\sigma_l^k - \widehat{\sigma}_l^k)^2}}{ \sum_{k=1}^K\sum_{l=1}^L \sigma_l^k },
\end{equation}
where $\sigma_l^k$ and $\widehat{\sigma}_l^k$ denote the ground truth speed and the estimated traffic state in lattice cell $l$ at time $k$, respectively. 
The stop-and-go-dynamics are well captured at all three penetration levels, but the RMSE and $\epsilon_{\sigma}^{\mathrm{rel}}$ values indicate improvement with increased penetration levels.
}

\subsection{Probe Vehicle Distribution}
\begin{figure}[h!]
	\centering
	\subfloat[][ Ground truth]{
		\includegraphics[width=0.45\textwidth]{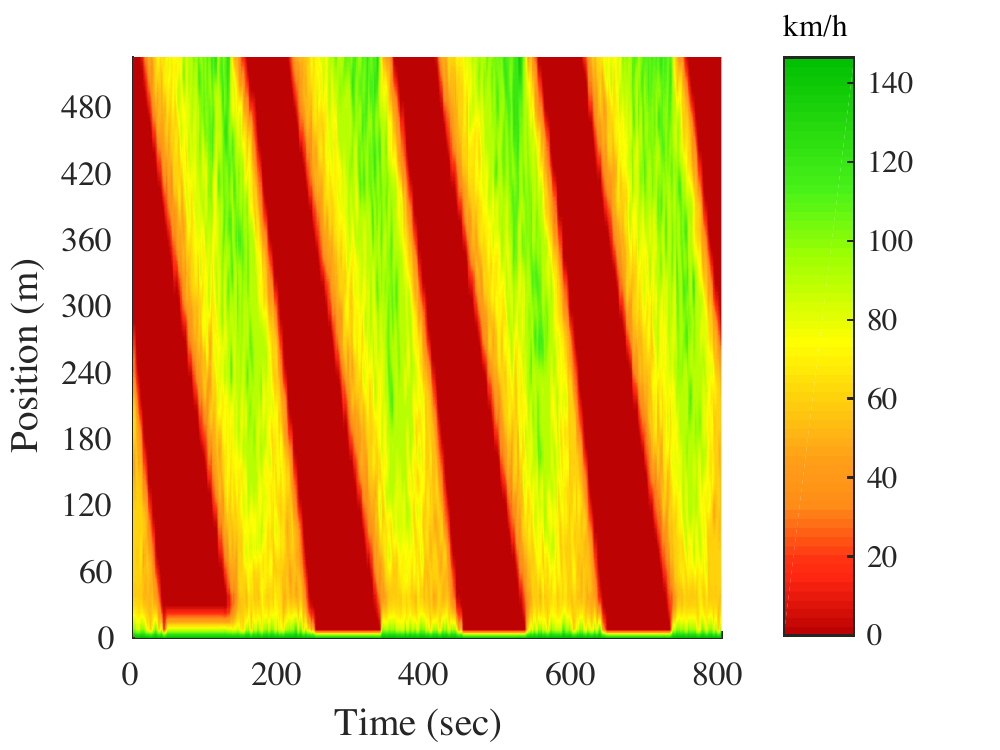}
		\label{fig:real1a}} 
	
	\subfloat[][Probe penetration = 10, $\epsilon_{\sigma}^{\mathrm{rel}}=0.1270$, RMSE = 18.69 kmph]{
		\includegraphics[width=0.45\textwidth]{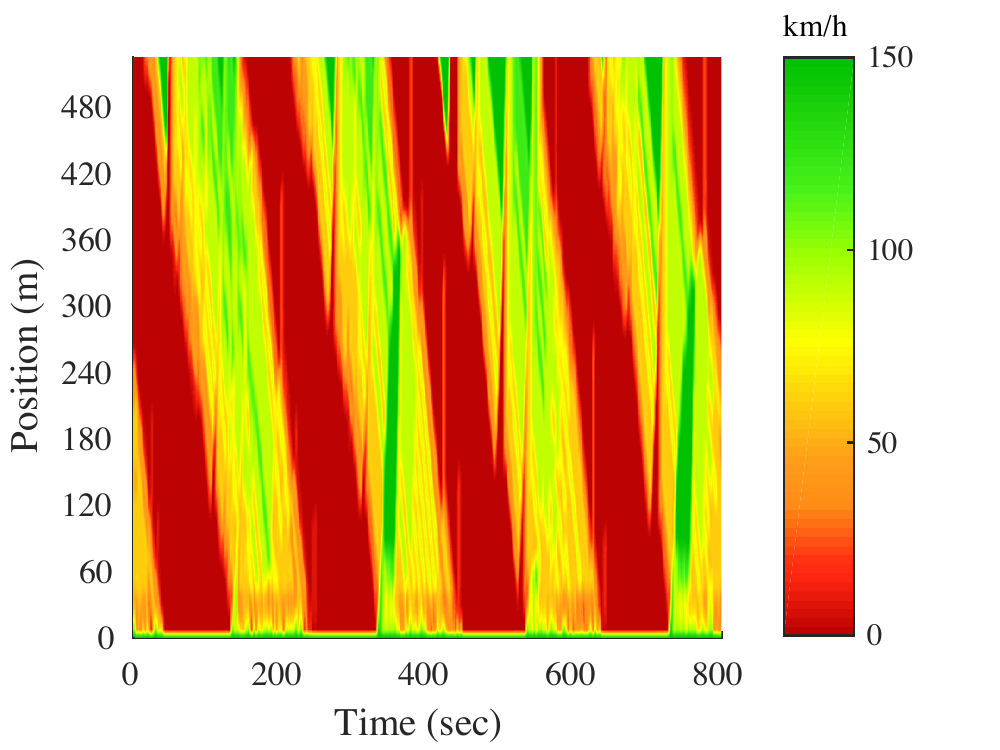}
		\label{fig:real1b}}

	\caption{Speed maps for signalized segment scenario.}
	\label{fig:real1}
\end{figure}

Accuracy of the estimation problem not only depends on penetration rates of probe vehicles, but also on how the probes are distributed in the sample.  In this section, we perform simulation experiments to investigate the impact of changing the spatial distribution of vehicles in a sample of probes used for reconstruction.  In this example, we simulate a 500m long arterial road segment with a fixed signal timing plan at the downstream end.  Four 200-second signal cycles are simulated with a green signal indication of 100sec and red signal indication of 100sec in each cycle.  Fig. \ref{fig:real1a} illustrates the resulting speed map and Fig. \ref{fig:real1b} illustrates a speed map that was reconstructed with a 10\% probe penetration (for reference).
 
\begin{figure}[ht!]
	\centering
	\resizebox{0.4\textwidth}{!}{%
		\includegraphics{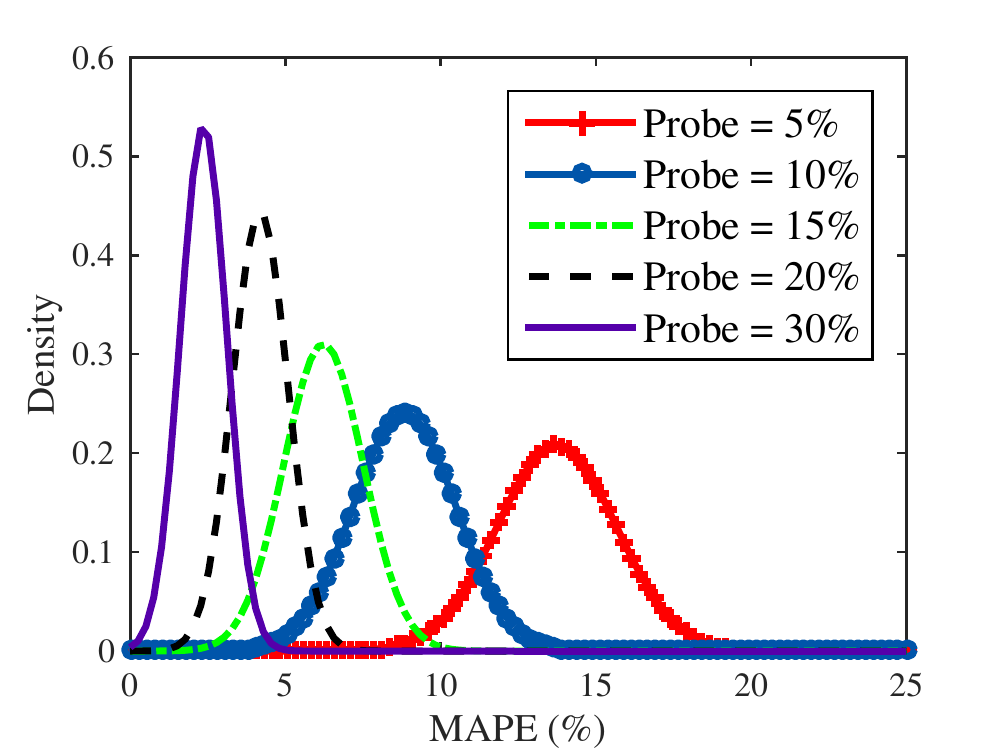}}
	\caption{PDF of the Mean Absolute Percent Error in travel time at different probe levels }
	\label{fig:pdf}
\end{figure}

To investigate the impact of probe distribution in the sample (spatially), 100 simulations were conducted for 5\%, 10\%, 20\%, and 30\% penetration rates and the speed map was reconstructed using the proposed MRF approach.  Fig. \ref{fig:pdf} depicts the frequencies of the MAPEs for each of the penetration rates.  

The mean value of the MAPE for a probe penetration level of $5\%$ is around $13\%$, but the high variability observed implies that travel time errors can be significantly higher under varying probe vehicle distributions. As the number of probe vehicles increases, this variability in the error decreases.

\subsection{Model Testing against Benchmark Algorithm}
\begin{figure}[ht!]%
	\centering
	
	\resizebox{0.5\textwidth}{!}{%
		\includegraphics{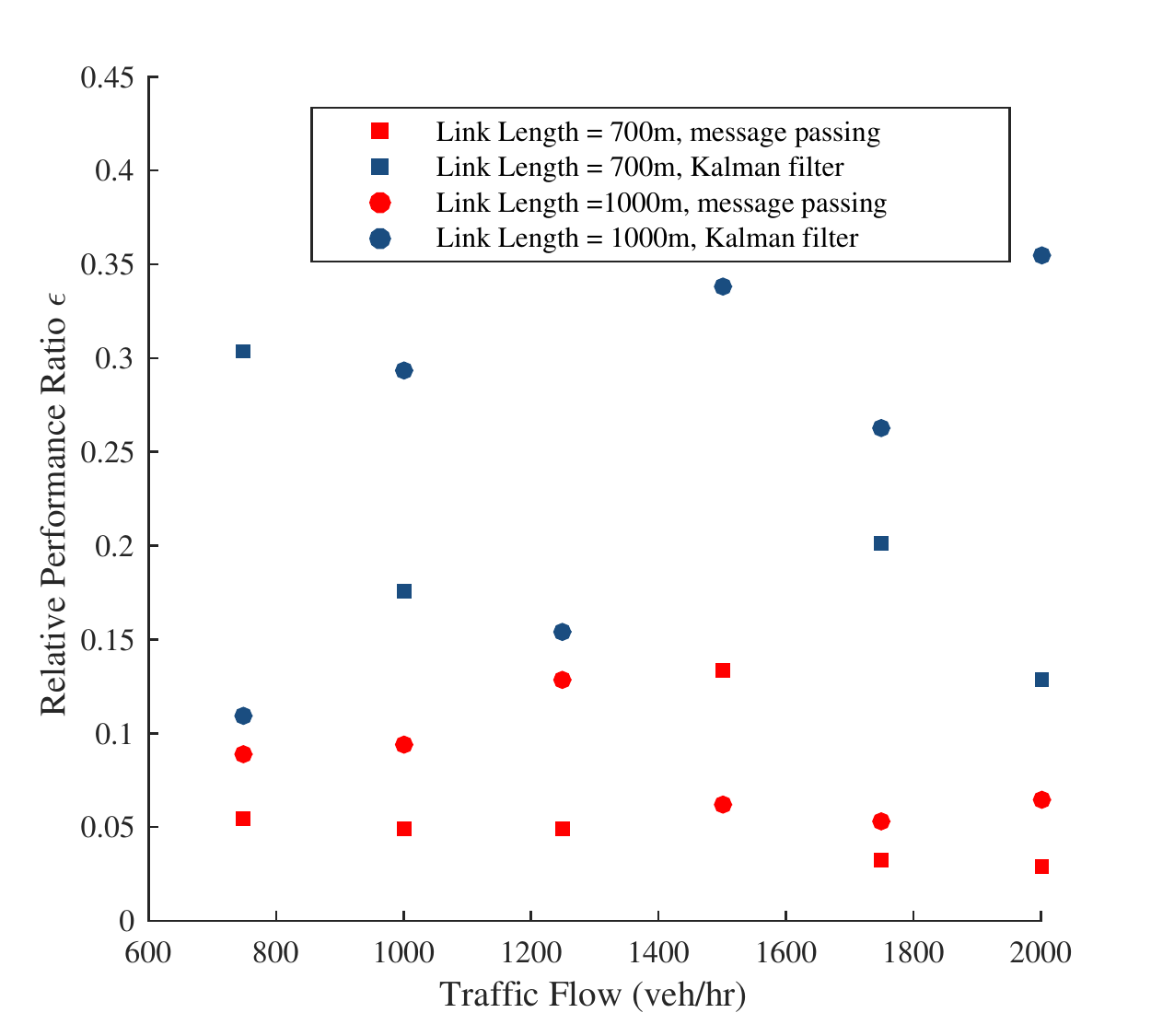}}
	
	\caption{Comparison with Benchmark Model }%
	\label{fig:2}%
\end{figure}

In this section, we next compare the proposed reconstruction technique against a Kalman filter, which is the most widely used approach for TSE, see \cite{seo2017traffic} for a recent survey.  For this purpose, we employ a recent implementation in  \cite{bekiaris2016highway} for the Kalman filter. Most of these approaches aim to estimate traffic densities, which can be extracted from proposed approach via the cell occupancies: let $\rho_l^k$ denote the traffic density in cell $l$ at time $k$,
\begin{equation}
\rho_l^k = \left\{
\begin{array}{ll} 
0 & \mbox{if } \sigma_l^k = -1 \\
\Delta l^{-1} & \mbox{otherwise}
\end{array}, \right.
\end{equation}
where $\Delta l$ is the cell size.  To compare estimation accuracy, we use the relative performance ratio $\epsilon_{\rho}^{\mathrm{rel}}$ (proposed in \cite{bekiaris2016highway}):
\begin{equation}
\epsilon_{\rho} = \frac{\sqrt{KL \sum_{k=1}^{K} \sum_{l=1}^{L} \big( \rho_l^k-\widehat{\rho}_l^k \big)^2 }}{ \sum_{k=1}^{K} \sum_{l=1}^{L}\rho_l(k)},
\end{equation}
where as before $\rho_l^k$ and $\boldsymbol{\rho}_l^k$ are the ground truth and estimated traffic densities at time $k$ in cell $l$.  The testing is done using different ground truth traffic conditions on a signalized roadway under various road section lengths and various inflow rates.  For both cases, we utilize a 15\% penetration rate.  As observed in Fig. \ref{fig:2}, the relative performance ratios achieved using the MRF approach are lower than those of the Kalman filter for all test cases, indicating higher accuracies.

\section{Conclusion}

A methodology is presented for TSE that combines mesoscopic traffic modeling with the statistical power of probabilistic graphical models.  A Markov random fields (MRF) approach using a factor graph representation of the dynamics is proposed for purposes of statistical learning when limited data is available.  The main insights are, first, that the random field representation and, more importantly, the factors, follow immediately from the traffic dynamics.  This implies that the estimated traffic states respect all traffic flow laws established by the traffic dynamics. Second, we represent the dynamics by a sequence of random fields, one per time step, which results in factor graphs that are trees.  Consequently, one can obtain the marginal distributions exactly via message passing techniques with a single forward and single backward pass (in each time step).  In each time step the inference problem is solved only once, which means that the proposed approach is amenable to online implementation.

Our experiments suggest that the proposed approach is capable of outperforming the standard techniques, which employ Kalman filters for TSE.  The main challenge in the present approach and all approaches that aim to estimate traffic conditions using Lagrangian models and measurements is the absence of knowledge about the number of vehicles in the system at any time step.  While several recent papers have addressed this problem, e.g., \cite{zheng2017estimating}, a TSE approach that combines estimation of volumes and vehicle speeds appears to continue to be an open problem. This could be a good avenue for future research.  Another possible extension to the present approach is one that includes multi-lane traffic.


%



\section*{Acknowledgment}
This work was funded in part by the C2SMART Center, a Tier 1 USDOT University Transportation Center and in part by the New York University Abu Dhabi Research Enhancement Fund.

\ifCLASSOPTIONcaptionsoff
  \newpage
\fi



\bibliographystyle{IEEEtran}
\bibliography{sample}

\begin{thebibliography}{10}
\providecommand{\url}[1]{#1}
\csname url@samestyle\endcsname
\providecommand{\newblock}{\relax}
\providecommand{\bibinfo}[2]{#2}
\providecommand{\BIBentrySTDinterwordspacing}{\spaceskip=0pt\relax}
\providecommand{\BIBentryALTinterwordstretchfactor}{4}
\providecommand{\BIBentryALTinterwordspacing}{\spaceskip=\fontdimen2\font plus
\BIBentryALTinterwordstretchfactor\fontdimen3\font minus
  \fontdimen4\font\relax}
\providecommand{\BIBforeignlanguage}[2]{{%
\expandafter\ifx\csname l@#1\endcsname\relax
\typeout{** WARNING: IEEEtran.bst: No hyphenation pattern has been}%
\typeout{** loaded for the language `#1'. Using the pattern for}%
\typeout{** the default language instead.}%
\else
\language=\csname l@#1\endcsname
\fi
#2}}
\providecommand{\BIBdecl}{\relax}
\BIBdecl

\bibitem{yuan2012real}
Y.~Yuan, J.~Van~Lint, E.~Wilson, F.~van Wageningen-Kessels, and S.~Hoogendoorn,
  ``Real-time {L}agrangian traffic state estimator for freeways,'' \emph{IEEE
  Transactions on Intelligent Transportation Systems}, vol.~13, no.~1, pp.
  59--70, 2012.

\bibitem{seo2015probe}
T.~Seo and T.~Kusakabe, ``Probe vehicle-based traffic state estimation method
  with spacing information and conservation law,'' \emph{Transportation
  Research Part C: Emerging Technologies}, vol.~59, pp. 391--403, 2015.

\bibitem{hofleitner2012learning}
A.~Hofleitner, R.~Herring, P.~Abbeel, and A.~Bayen, ``Learning the dynamics of
  arterial traffic from probe data using a dynamic {B}ayesian network,''
  \emph{IEEE Transactions on Intelligent Transportation Systems}, vol.~13,
  no.~4, pp. 1679--1693, 2012.

\bibitem{herrera2010incorporation}
J.~Herrera and A.~Bayen, ``Incorporation of {L}agrangian measurements in
  freeway traffic state estimation,'' \emph{Transportation Research Part B:
  Methodological}, vol.~44, no.~4, pp. 460--481, 2010.

\bibitem{jabari2012stochastic}
S.~Jabari and H.~Liu, ``A stochastic model of traffic flow: {Theoretical}
  foundations,'' \emph{Transportation Research Part B: Methodological},
  vol.~46, no.~1, pp. 156--174, 2012.

\bibitem{jabari2013stochastic}
------, ``A stochastic model of traffic flow: {Gaussian} approximation and
  estimation,'' \emph{Transportation Research Part B: Methodological}, vol.~47,
  pp. 15--41, 2013.

\bibitem{deng2013traffic}
W.~Deng, H.~Lei, and X.~Zhou, ``Traffic state estimation and uncertainty
  quantification based on heterogeneous data sources: {A} three detector
  approach,'' \emph{Transportation Research Part B: Methodological}, vol.~57,
  pp. 132--157, 2013.

\bibitem{work2008ensemble}
D.~Work, O.~Tossavainen, S.~Blandin, A.~Bayen, T.~Iwuchukwu, and K.~Tracton,
  ``An ensemble {K}alman filtering approach to highway traffic estimation using
  {GPS} enabled mobile devices,'' in \emph{47th IEEE Conference on Decision and
  Control (CDC)}.\hskip 1em plus 0.5em minus 0.4em\relax IEEE, 2008, pp.
  5062--5068.

\bibitem{hellinga2008decomposing}
B.~Hellinga, P.~Izadpanah, H.~Takada, and L.~Fu, ``Decomposing travel times
  measured by probe-based traffic monitoring systems to individual road
  segments,'' \emph{Transportation Research Part C: Emerging Technologies},
  vol.~16, no.~6, pp. 768--782, 2008.

\bibitem{chen2012prediction}
H.~Chen and H.~Rakha, ``Prediction of dynamic freeway travel times based on
  vehicle trajectory construction,'' in \emph{15th International IEEE
  Conference on Intelligent Transportation Systems (ITSC)}.\hskip 1em plus
  0.5em minus 0.4em\relax IEEE, 2012, pp. 576--581.

\bibitem{hunter2012large}
T.~Hunter, T.~Das, M.~Zaharia, P.~Abbeel, and A.~Bayen, ``Large scale
  estimation in cyberphysical systems using streaming data: {A} case study with
  smartphone traces,'' \emph{arXiv preprint arXiv:1212.3393}, pp. 1--13, 2012.

\bibitem{jenelius2017urban}
E.~Jenelius and H.~Koutsopoulos, ``Urban network travel time prediction based
  on a probabilistic principal component analysis model of probe data,''
  \emph{IEEE Transactions on Intelligent Transportation Systems}, pp. 436--445,
  2017.

\bibitem{dilip2017sparse}
D.~Dilip, N.~Freris, and S.~Jabari, ``Sparse estimation of travel time
  distributions using {G}amma kernels,'' \emph{Proceedings of the 96th Annual
  Meeting of the Transportation Research Board (No. 17-02971)}, 2017.

\bibitem{jabariGenGammaKernels}
\BIBentryALTinterwordspacing
S.~Jabari, N.~Freris, and D.~Dilip, ``Sparse travel time estimation from
  streaming data,'' \emph{Transportation Science}, vol. (in press), 2019.
  [Online]. Available: \url{https://arxiv.org/abs/1804.08130}
\BIBentrySTDinterwordspacing

\bibitem{herring2010estimating}
R.~Herring, A.~Hofleitner, P.~Abbeel, and A.~Bayen, ``Estimating arterial
  traffic conditions using sparse probe data,'' in \emph{2010 13th
  International IEEE Conference on Intelligent Transportation Systems
  (ITSC)}.\hskip 1em plus 0.5em minus 0.4em\relax IEEE, 2010, pp. 929--936.

\bibitem{hofleitner2012arterial}
A.~Hofleitner, R.~Herring, and A.~Bayen, ``Arterial travel time forecast with
  streaming data: {A} hybrid approach of flow modeling and machine learning,''
  \emph{Transportation Research Part B: Methodological}, vol.~46, no.~9, pp.
  1097--1122, 2012.

\bibitem{papathanasopoulou2015towards}
V.~Papathanasopoulou and C.~Antoniou, ``Towards data-driven car-following
  models,'' \emph{Transportation Research Part C: Emerging Technologies},
  vol.~55, pp. 496--509, 2015.

\bibitem{furtlehner2007belief}
C.~Furtlehner, J.~Lasgouttes, and A.~de~La~Fortelle, ``A belief propagation
  approach to traffic prediction using probe vehicles,'' in \emph{IEEE
  Intelligent Transportation Systems Conference (ITSC)}.\hskip 1em plus 0.5em
  minus 0.4em\relax IEEE, 2007, pp. 1022--1027.

\bibitem{bekiaris2016highway}
N.~Bekiaris-Liberis, C.~Roncoli, and M.~Papageorgiou, ``Highway traffic state
  estimation with mixed connected and conventional vehicles,'' \emph{IEEE
  Transactions on Intelligent Transportation Systems}, vol.~17, no.~12, pp.
  3484--3497, 2016.

\bibitem{fountoulakis2017highway}
M.~Fountoulakis, N.~Bekiaris-Liberis, C.~Roncoli, I.~Papamichail, and
  M.~Papageorgiou, ``Highway traffic state estimation with mixed connected and
  conventional vehicles: Microscopic simulation-based testing,''
  \emph{Transportation Research Part C: Emerging Technologies}, vol.~78, pp.
  13--33, 2017.

\bibitem{vandenberghe2012feasibility}
W.~Vandenberghe, E.~Vanhauwaert, S.~Verbrugge, I.~Moerman, and P.~Demeester,
  ``Feasibility of expanding traffic monitoring systems with floating car data
  technology,'' \emph{IET Intelligent Transport Systems}, vol.~6, no.~4, pp.
  347--354, 2012.

\bibitem{jabari2016Sensor}
S.~Jabari and L.~Wynter, ``Sensor placement with time-to-detection
  guarantees,'' \emph{EURO Journal on Transportation and Logistics}, vol.~5,
  no.~4, pp. 415--433, 2016.

\bibitem{mazare2012trade}
P.~Mazar{\'e}, O.~Tossavainen, A.~Bayen, and D.~Work, ``Trade-offs between
  inductive loops and {GPS} probe vehicles for travel time estimation: {A}
  {M}obile {C}entury case study,'' \emph{Proceedings of the 91st Annual Meeting
  of the Transportation Research Board (No. 12-2746)}, 2012.

\bibitem{kerner2013traffic}
B.~S. Kerner, H.~Rehborn, R.-P. Sch{\"a}fer, S.~L. Klenov, J.~Palmer,
  S.~Lorkowski, and N.~Witte, ``Traffic dynamics in empirical probe vehicle
  data studied with three-phase theory: Spatiotemporal reconstruction of
  traffic phases and generation of jam warning messages,'' \emph{Physica a:
  statistical mechanics and its applications}, vol. 392, no.~1, pp. 221--251,
  2013.

\bibitem{hiribarren2014real}
G.~Hiribarren and J.~Herrera, ``Real time traffic states estimation on
  arterials based on trajectory data,'' \emph{Transportation Research Part B:
  Methodological}, vol.~69, pp. 19--30, 2014.

\bibitem{ban2009delay}
X.~Ban, R.~Herring, P.~Hao, and A.~Bayen, ``Delay pattern estimation for
  signalized intersections using sampled travel times,'' \emph{Transportation
  Research Record: Journal of the Transportation Research Board}, no. 2130, pp.
  109--119, 2009.

\bibitem{ban2011real}
X.~Ban, P.~Hao, and Z.~Sun, ``Real time queue length estimation for signalized
  intersections using travel times from mobile sensors,'' \emph{Transportation
  Research Part C: Emerging Technologies}, vol.~19, no.~6, pp. 1133--1156,
  2011.

\bibitem{zheng2018traffic}
F.~Zheng, S.~Jabari, H.~Liu, and D.~Lin, ``Traffic state estimation using
  stochastic {L}agrangian dynamics,'' \emph{Transportation Research Part B:
  Methodological}, vol. 115, pp. 143--165, 2018.

\bibitem{dilip2018vehicle}
D.~Dilip and S.~Jabari, ``Vehicle trajectory reconstruction using conditional
  random fields,'' \emph{Proceedings of the 97th Annual Meeting of the
  Transportation Research Board (No. 18-03053)}, 2018.

\bibitem{kim2014comparing}
S.~Kim and B.~Coifman, ``Comparing {INRIX} speed data against concurrent loop
  detector stations over several months,'' \emph{Transportation Research Part
  C: Emerging Technologies}, vol.~49, pp. 59--72, 2014.

\bibitem{bar2007evaluation}
H.~Bar-Gera, ``Evaluation of a cellular phone-based system for measurements of
  traffic speeds and travel times: {A} case study from israel,''
  \emph{Transportation Research Part C: Emerging Technologies}, vol.~15, no.~6,
  pp. 380--391, 2007.

\bibitem{sugiyama2008traffic}
Y.~Sugiyama, M.~Fukui, M.~Kikuchi, K.~Hasebe, A.~Nakayama, K.~Nishinari,
  S.~Tadaki, and S.~Yukawa, ``Traffic jams without bottlenecks—experimental
  evidence for the physical mechanism of the formation of a jam,'' \emph{New
  Journal of Physics}, vol.~10, no.~3, p. 033001, 2008.

\bibitem{sopasakis2006stochastic}
A.~Sopasakis and M.~Katsoulakis, ``Stochastic modeling and simulation of
  traffic flow: {A}symmetric single exclusion process with {A}rrhenius
  look-ahead dynamics,'' \emph{SIAM Journal on Applied Mathematics}, vol.~66,
  no.~3, pp. 921--944, 2006.

\bibitem{larraga2005cellular}
M.~L{\'a}rraga, J.~Del~Rio, and L.~Alvarez-Lcaza, ``Cellular automata for
  one-lane traffic flow modeling,'' \emph{Transportation Research Part C:
  Emerging Technologies}, vol.~13, no.~1, pp. 63--74, 2005.

\bibitem{wagner1997realistic}
P.~Wagner, K.~Nagel, and D.~E. Wolf, ``Realistic multi-lane traffic rules for
  cellular automata,'' \emph{Physica A: Statistical Mechanics and its
  Applications}, vol. 234, no. 3-4, pp. 687--698, 1997.

\bibitem{li2006realistic}
X.-G. Li, B.~Jia, Z.-Y. Gao, and R.~Jiang, ``A realistic two-lane cellular
  automata traffic model considering aggressive lane-changing behavior of fast
  vehicle,'' \emph{Physica A: Statistical Mechanics and its Applications}, vol.
  367, pp. 479--486, 2006.

\bibitem{emmerich1997improved}
H.~Emmerich and E.~Rank, ``An improved cellular automaton model for traffic
  flow simulation,'' \emph{Physica A: Statistical Mechanics and its
  Applications}, vol. 234, no. 3-4, pp. 676--686, 1997.

\bibitem{jabari2016node}
S.~Jabari, ``Node modeling for congested urban road networks,''
  \emph{Transportation Research Part B: Methodological}, vol.~91, pp. 229--249,
  2016.

\bibitem{schadschneider2002traffic}
A.~Schadschneider, ``Traffic flow: a statistical physics point of view,''
  \emph{Physica A: Statistical Mechanics and its Applications}, vol. 313, no.
  1-2, pp. 153--187, 2002.

\bibitem{tian2009synchronized}
J.-f. Tian, B.~Jia, X.-g. Li, R.~Jiang, X.-m. Zhao, and Z.-y. Gao,
  ``Synchronized traffic flow simulating with cellular automata model,''
  \emph{Physica A: Statistical Mechanics and its Applications}, vol. 388,
  no.~23, pp. 4827--4837, 2009.

\bibitem{zheng2017estimating}
J.~Zheng and H.~Liu, ``Estimating traffic volumes for signalized intersections
  using connected vehicle data,'' \emph{Transportation Research Part C},
  vol.~79, pp. 347--362, 2017.

\bibitem{bishop2006pattern}
C.~Bishop, \emph{Pattern recognition and machine learning}.\hskip 1em plus
  0.5em minus 0.4em\relax springer, 2006.

\bibitem{koller2009probabilistic}
D.~Koller and N.~Friedman, \emph{Probabilistic graphical models: {P}rinciples
  and techniques}.\hskip 1em plus 0.5em minus 0.4em\relax MIT press, 2009.

\bibitem{lu2007freeway}
X.-Y. Lu and A.~Skabardonis, ``Freeway traffic shockwave analysis: exploring
  the ngsim trajectory data,'' in \emph{86th Annual Meeting of the
  Transportation Research Board, Washington, DC}, 2007.

\bibitem{seo2017traffic}
T.~Seo, A.~Bayen, T.~Kusakabe, and Y.~Asakura, ``Traffic state estimation on
  highway: {A} comprehensive survey,'' \emph{Annual Reviews in Control},
  vol.~43, pp. 128--151, 2017.

\end{thebibliography}
%



%

\begin{IEEEbiography}[{\includegraphics[width=1in,height=1.25in,clip,keepaspectratio]{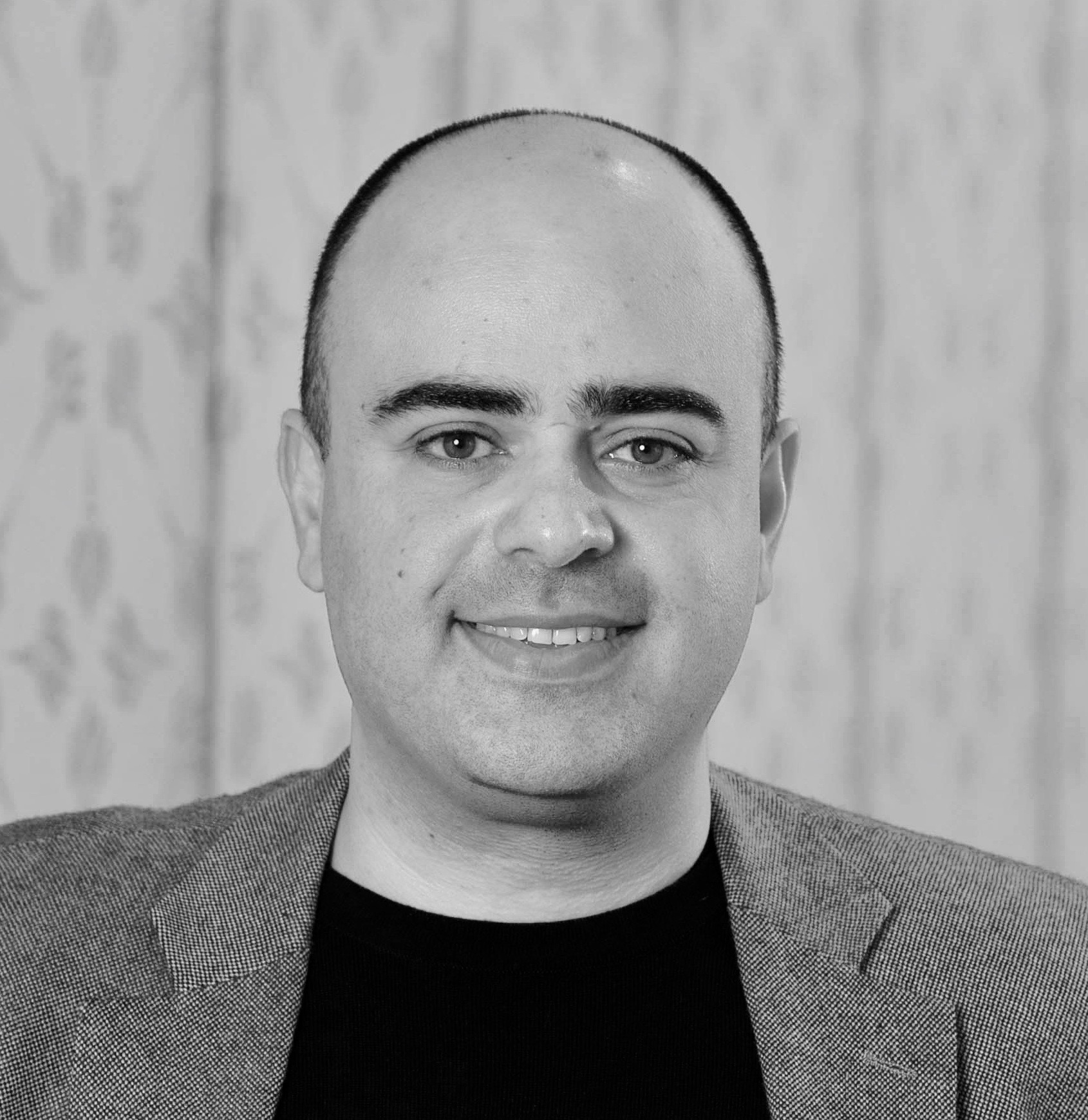}}]{Saif Eddin Jabari}
	is an Assistant Professor of Civil and Urban Engineering at New York University Abu Dhabi, U.A.E. and Global Network Assistant Professor of Civil and Urban Engineering at the Tandon School of Engineering, New York University, NY, U.S.A.  He received his B.Sc. degree in Civil Engineering from the University of Jordan and his M.S. and Ph.D. degrees in Civil Engineering from the University of Minnesota Twin Cities in 2009 and 2012, respectively. Prior to joining New York University, he was a Post-Doctoral Researcher in the Math Department at the IBM Watson Research Center in Yorktown Heights, New York.  His research interests lie at the intersection of theoretical traffic flow modeling and statistical learning, with applications in traffic operations, including traffic state estimation, incident detection and localization, and real-time traffic control.
\end{IEEEbiography}

\begin{IEEEbiography}[{\includegraphics[width=1in,height=1.25in,clip,keepaspectratio]{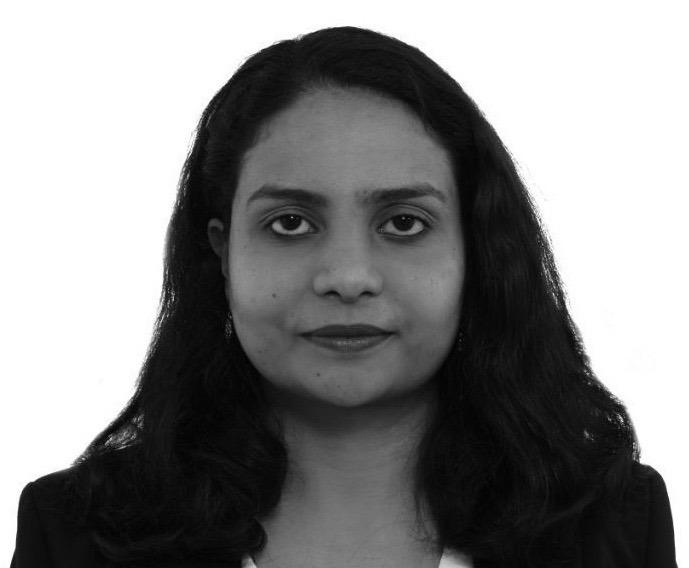}}]{Deepthi Mary Dilip}
is an Assistant Professor at BITS Pilani, Dubai Campus. She received her M.Tech and Ph.D. degrees from National Institute of Technology, Calicut and Indian Institute of Science, Bangalore in 2009 and 2015, respectively. She held a Post-Doctoral Associate position with New York University Abu Dhabi prior to joining BITS Pilani. Her research interests include probabilistic modeling of transportation systems and reliability analysis.
\end{IEEEbiography}

\begin{IEEEbiography}[{\includegraphics[width=1in,height=1.25in,clip,keepaspectratio]{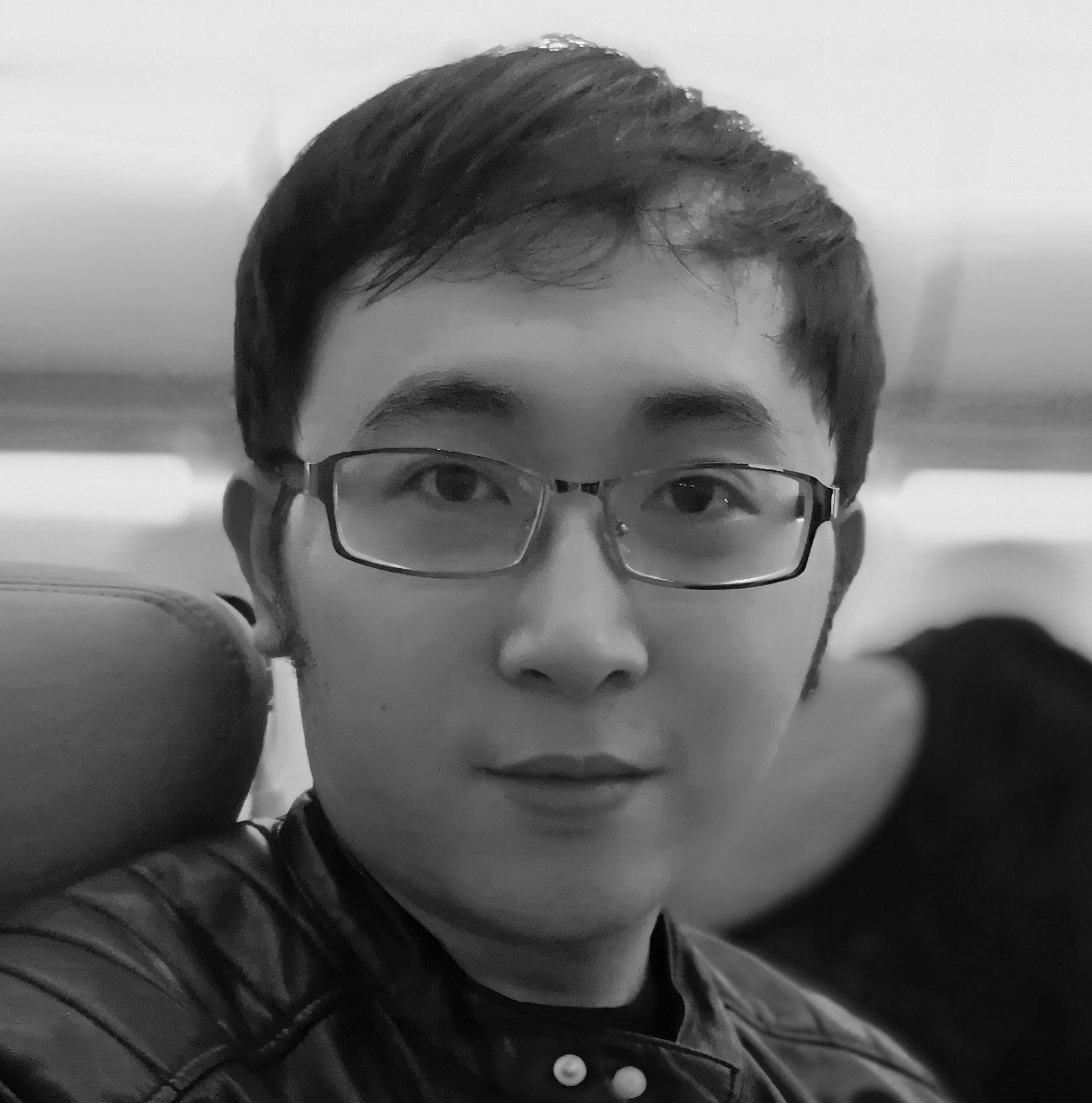}}]{DianChao Lin}
received his B.Sc. and M.Sc. degrees in Traffic Engineering and Traffic Information Engineering \& Control from Tongji University, Shanghai, China, in 2013 and 2016, respectively. He is currently pursuing his Ph.D. degree in Transportation Planning \& Engineering at New York University, NY, U.S.A. His research interests include traffic state estimation, applications of game theory to automated vehicles, and multi-modal traffic flow in urban networks.
\end{IEEEbiography}

\begin{IEEEbiography}[{\includegraphics[width=1in,height=1.25in,clip,keepaspectratio]{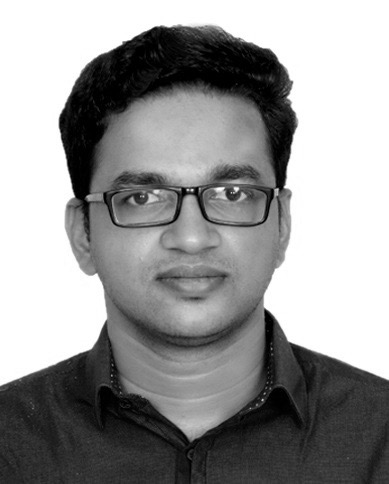}}]{Bilal Thonnam Thodi}
	received his B.Tech. degree in Civil Engineering from Cochin University of Science and Technology in 2014 and M.Tech. degree in Transportation Engineering from The Indian Institute of Technology Madras (IITM) in 2017. He is currently pursuing his Ph.D. degree in Transportation Planning \& Engineering at New York University, NY, U.S.A. His research interests include applications of artificial intelligence and machine learning techniques to problems in traffic operations.
\end{IEEEbiography}




\end{document}